\documentclass[letterpaper,journal,oneside]{IEEEtran}
\usepackage{amsmath,amsfonts}
\usepackage{amssymb}
\usepackage{algorithmic}
\usepackage{algorithm}
\usepackage{array}
\usepackage[caption=false,font=normalsize,labelfont=sf,textfont=sf]{subfig}
\usepackage{textcomp}
\usepackage{stfloats}
\usepackage{url}
\usepackage{verbatim}
\usepackage{graphicx}
\usepackage{booktabs}
\usepackage{hyperref}
\usepackage{newunicodechar}
\newunicodechar{′}{\ensuremath{'} }
\newunicodechar{″}{\ensuremath{"{}} }
\usepackage{cite}
\usepackage{authblk}
\newcommand{\vn}[1]{\mathrm{#1}}

\begin{document}

\title{Automated Palynological Analysis System: Integrating Deep Metric Learning, Detection and Classification in Bright Field Microscopy}

\author{J.~Staforelli-Vivanco, R.~Jofré, P.~Coelho, I.~Sanhueza, L.~Viafora, C.~Toro, J.~Troncoso, M.~Rondanelli-Reyes, I.~Lamas, Andy~Banegas-Medina, Isis-Yelena~Montes, B.~Muñoz-Cepeda, V.~Salamanca-Levi, M.~González-Ortiz, and E.~Vera%
\thanks{J. Staforelli-Vivanco (jstaforelli@udec.cl), R. Jofré, B. Muñoz, and V. Salamanca are with Departamento de Física, Facultad de Ciencias Físicas y Matemáticas, Universidad de Concepción, Concepción, Chile}%
\thanks{P. Coelho, I. Sanhueza and L. Viafora are with Facultad de Ingeniería, Universidad San Sebastián, Concepción, 4080871, Chile}%
\thanks{C. Toro is with Facultad de Ingeniería, Universidad Andres Bello, Talcahuano, CCP-THNO 7100 Chile}%
\thanks{J. Troncoso is with Facultad Universidad Arturo Prat, Sede Victoria, Chile}%
\thanks{M. Rondanelli-Reyes and I. Lamas are with Laboratory of Palynology and Plant Ecology, School of Sciences and Technologies, University of Concepción, Los Angeles Campus}%
\thanks{Andy Banegas-Medina and I. Isis-Yelena Montes are with Laboratory of Biology, Department of Sciences, Danli Technological Campus, Universidad Nacional Autonoma de Honduras, Danli, Honduras}%
\thanks{M. González-Ortiz is with Laboratorio de Investigación Materno-Fetal (LIMaF), Departamento de Obstetricia y Ginecología, Facultad de Medicina, Universidad de Concepción, Concepción, Chile}%
\thanks{E. Vera is with School of Electrical Engineering, Pontificia Universidad Católica de Valparaíso, Valparaíso, Chile.}%
\thanks{Manuscript received Jun 08, 2026; revised Jun 15, 2026.}}

\markboth{Journal of \LaTeX\ Class Files,~Vol.~14, No.~8, May~2026}%
{Shell \MakeLowercase{\textit{et al.}}: A Sample Article Using IEEEtran.cls for IEEE Journals}


\maketitle

\begin{abstract}
Traditional melissopalynology is a time-consuming and subjective process, often taking 4-6 hours per sample. We present an automated, high-throughput microscopy system that integrates $H\infty$ robust mechanical control with advanced deep learning pipelines for the precise counting, classification, and morphological analysis of pollen grains from Bio Bio region in south central territory in Chile. Our system employs $U^{2}$-Net for salient object detection  and a DINOv2 Vision Transformer backbone trained via Deep Metric Learning for classification. By integrating Gradient-Weighted Attention, the model provides human-interpretable texture and diagnostic feature annotations. The system achieves a 95.8$\%$ classification recall  and at least 6x processing speedup compared to manual expert analysis.
\end{abstract}


\section{Introduction}

\IEEEPARstart{T}{he} precise identification of pollen grains is essential for honey certification and fraud prevention \cite{Machuca2022,Jofre2025,sevillano2020precise}. In Chile, authenticating endemic honeys—particularly from biodiversity hotspots like the Pitril sector ($37^{\circ} 47' 13.7''S$, $71^{\circ} 32' 11.9''W$)—is vital for economic valuation and biodiversity monitoring. However, traditional manual microscopy is labor-intensive, often requiring up to six hours per sample due to operator fatigue and low throughput. To overcome these bottlenecks, we propose an automated framework that integrates a custom motorized microscopy platform with state-of-the-art Vision Transformers (ViTs) \cite{oquab2023dinov2, chefer2021transformer,dosovitskiy2021image,touvron2022deit}.  Our system synergizes hardware stabilized by an $H\infty$ robust controller \cite{Doyle1988,Doyle1996} with an AI pipeline featuring $U^2$-Net for salient object detection and a DINOv2 backbone optimized via Deep Metric Learning (DML) \cite{qin2020u2net,caron2021emerging}. Unlike traditional Softmax classifiers that require retraining for new categories, our DML approach maps pollen to a 128-dimensional unit hypersphere ($f_\theta: \mathbb{R}^{H \times W} \rightarrow \mathbb{S}^{127}$) using Multi-Similarity Loss \cite{musgrave2020metric,roth2022nonisotropy,kim2023hier,wang2019multi}. This ensures that Euclidean distances directly reflect taxonomic dissimilarity, facilitating zero-shot generalization and few-shot learning. The architecture achieves a six-fold reduction in processing time and provides explainable diagnostic heatmaps via Gradient-Weighted Attention \cite{selvaraju2017grad}.  The reference database is meticulously curated to represent the ecological diversity of the Biobío Region \cite{Poulsen2023,García2020}. It includes three groups: endemic species (\textit{Quillaja saponaria}, \textit{Lithrea caustica}), native flora (\textit{Embothrium coccineum}, \textit{Acaena splendens}), and introduced species (\textit{Medicago sativa}, \textit{Trifolium pratense}, \textit{Cucurbita pepo}). By providing high-quality taxonomic data from South America, this work addresses geographic bias in global datasets \cite{Figueroa2023,Staforelli2026,Sanhueza2026} and enhances model interpretability through direct metric space calculations.  The remainder of this article is organized as follows: Section 2 presents the sample collection sites. Section 3 describes the automated hardware, and $U^2$-Net localization. Section 4 details the DML classification system, DINOv2 optimization, and normalized background preprocessing. Section 5 present performance metrics, interpretability maps and operationa efficiency. Finally conclusions are presented in Section 6.

\section{Study Area and Botanical Sample Collection}

\begin{figure}[htbp]
    \centering
\includegraphics[width=0.45\textwidth]{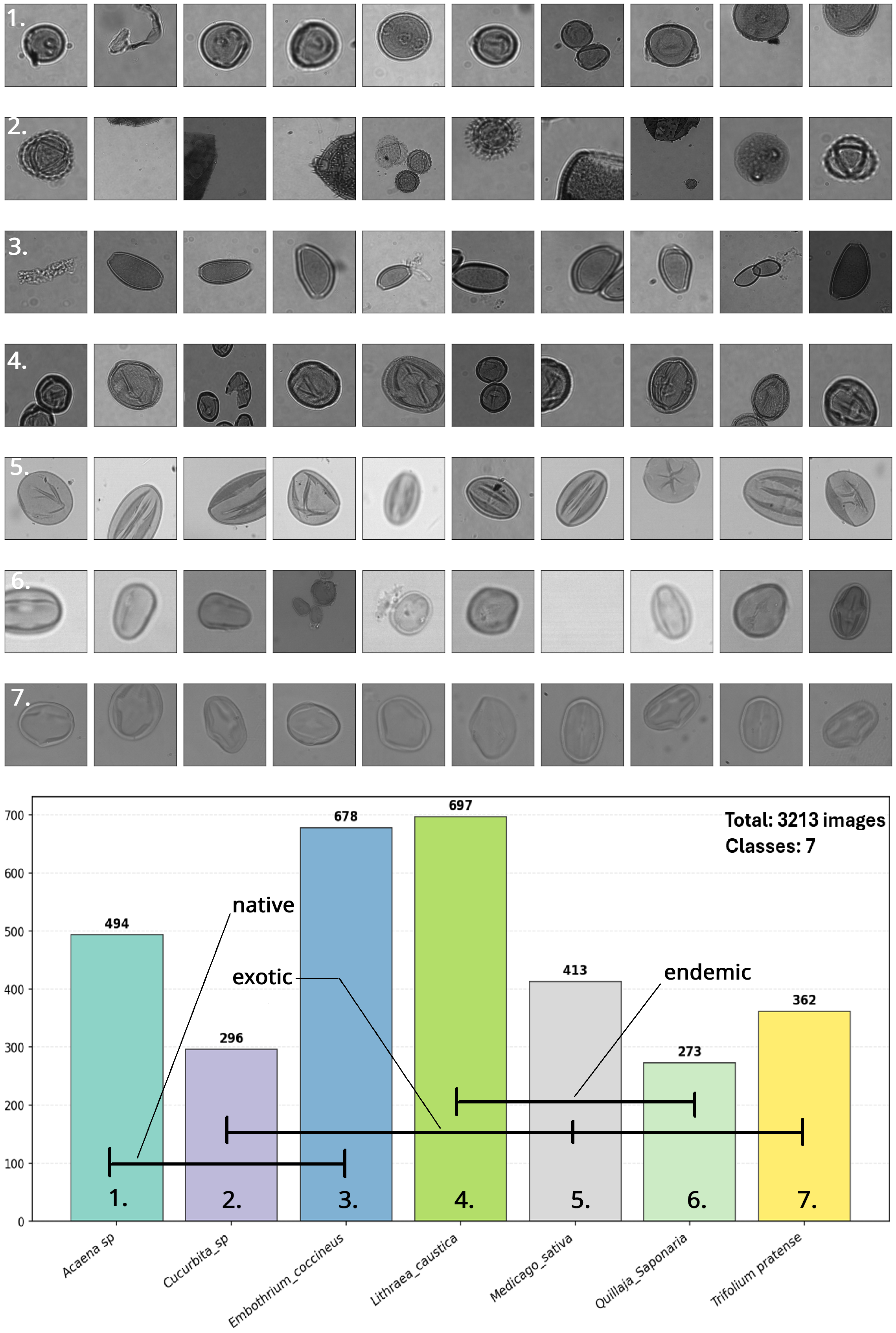}
    \caption{Sample grid per class and classes distribution. 1. \textit{Acaena splendens} (bidibid), 2. \textit{Cucurbita pepo} (squash), 3. \textit{Embothrium coccineum} (Chilean firebush), 4. \textit{Lithrea caustica }(litre tree), 5. \textit{ Medicago sativa} (alfalfa), 6. \textit{Quillaja saponaria} (soapbark) and 7. \textit{Trifolium pratense} (red clover). Connectors unite pollen classes by geobotany category: ``1,3": Native; ``2,5,7": Exotic and ``4,6": Endemic species. Upper number in bars: Images number from total.}
\label{classes}
\end{figure}

\begin{figure}[htbp]
    \centering
\includegraphics[width=0.45\textwidth]{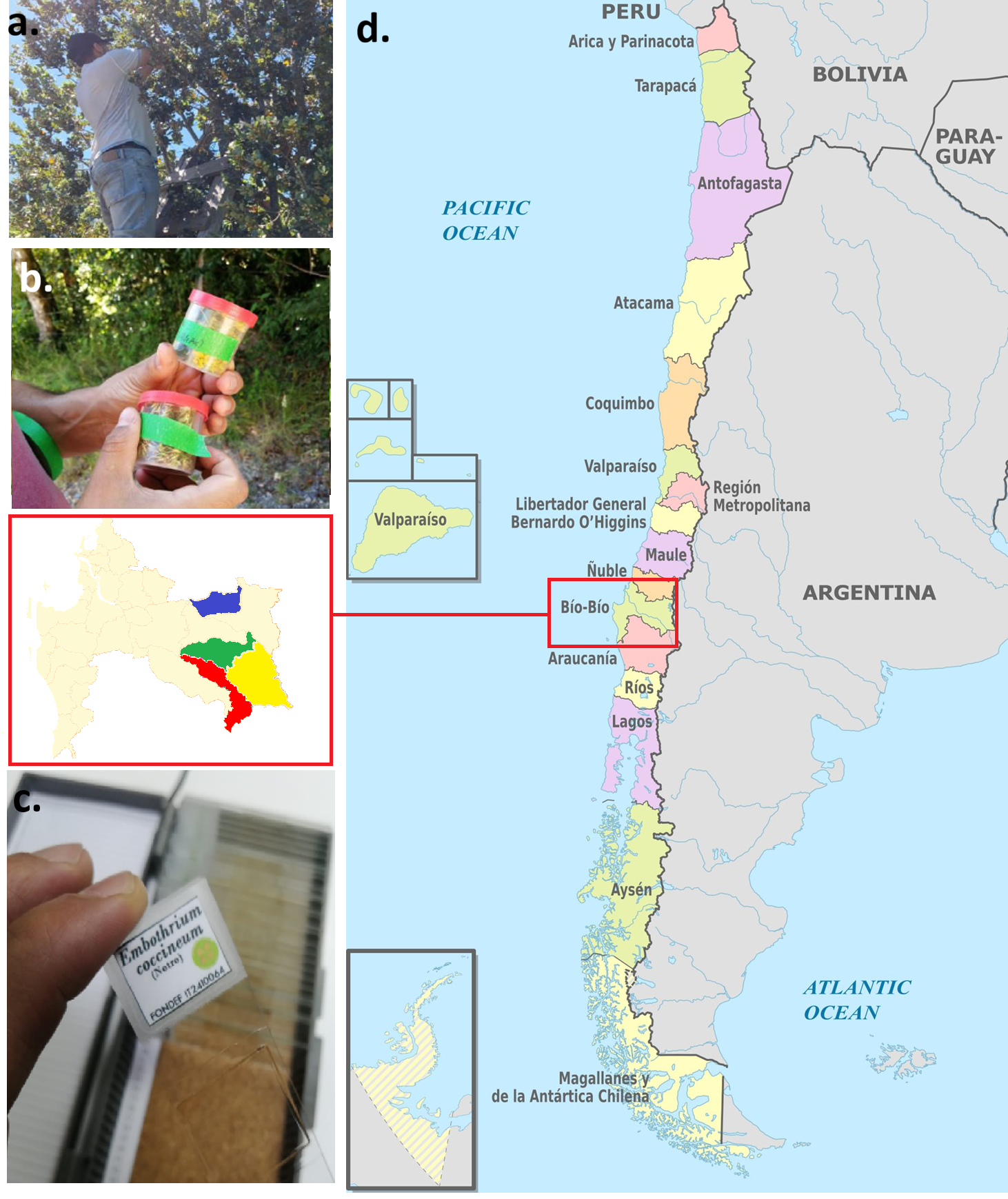}
    \caption{a-b. Botanical sample and reference pollen collection during the peak flowering season (Oct–Apr). c. Complete 7-entry palynotheca registry. d. Biobío region, Chile, with sampled communes color-coded: Quilaco (red), Santa Bárbara (green), Alto Biobío (yellow), and Yumbel (blue).}
    \label{map}
\end{figure}

To construct the database and reference palynotheca (Fig.~\ref{classes}), botanical samples were collected from four locations in the Biobío Region, Chile (Fig.~\ref{map}). These included the native flora-rich Pitril sector in Alto Biobío (37°47’13.7”S, 71°32’11.9”W) and a Yumbel site (37°02’02.1”S, 72°40’20.3”W) featuring 125 hectares of native reforestation (predominantly Quillaja saponaria). Additional zones included Santa Bárbara (37°40′00”S, 72°01′00”W) and Quilaco (37°45′00”S, 71°35′00”W). Across six campaigns, flowering plants within a 2 km radius of study apiaries were collected in triplicate. This allowed for taxonomic determination, chemical treatment for the palynotheca, and herborization. Voucher specimens were deposited at the Universidad de Concepción, Los Ángeles Campus.

Sample preparation and subsequent palynological analysis were conducted using the acetolysis method described by Faegri et al. (1989) \cite{louveaux1978methods, Faegri1989}. The procedure began with an acid hydrolysis to eliminate organic residues or biological wastes—preserving only the pollen grains due to their chemical resistance. For purification, a mixture of pure acetic anhydride and concentrated sulfuric acid was employed at a 9:1 ratio. This treatment removes the outer layers of the pollen grains, leaving only the exine (the outer shell) exposed, which facilitates morphological identification. Subsequently, the resulting concentrate was mounted using Hydromatrix \copyright mounting medium.

\section{Automated Microscopy and Detection}
\subsection{Hardware and Robust Control}
\begin{figure}[htbp]
    \centering
\includegraphics[width=0.45\textwidth]{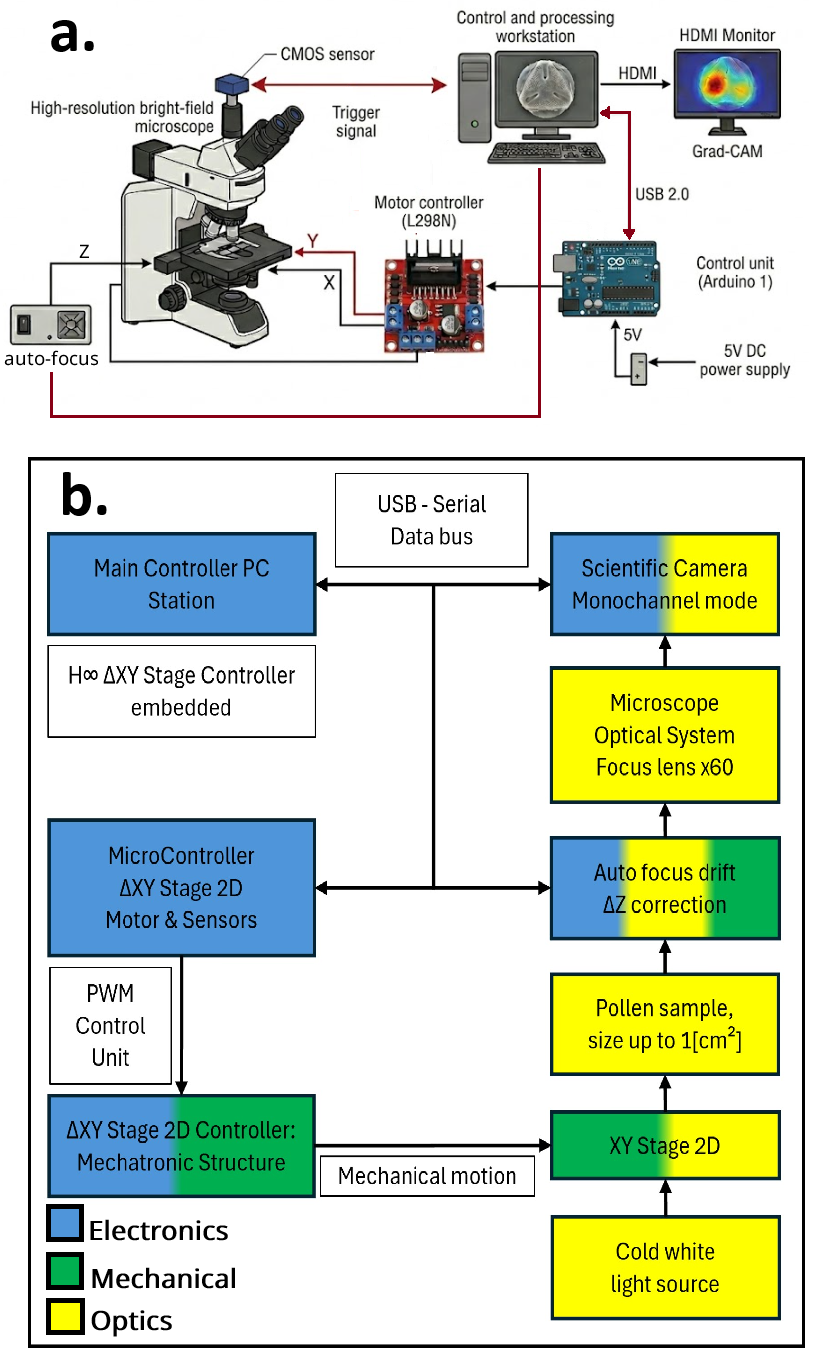}
    \caption{a. The automated system features a central workstation that coordinates an XY motorized stage and intelligent Z-axis autofocus for high-throughput scanning. Triggered blur-free CMOS images are instantly analyzed on-the-fly, classifying pollen grains and outputting interpretable Grad-CAM heatmaps. b. System flow diagram (blue: electronics, green: mechanical, yellow: optics). }
    \label{setup}
\end{figure}

The imaging setup is built around a compound optical microscope operating in brightfield mode, utilizing Köhler illumination to ensure a uniform field of view \cite{Sohn2006}. High-resolution image acquisition is performed with a digital CCD/CMOS sensor (minimum 1920x1080 pixels) using a 60x oil immersion objective lens from NIKON. The mechanical scanning platform of samples consists of an XY stage driven by DC motors and a piezoelectric C-Focus$^{TM}$ system for the Z-axis control \cite{madcity}. To overcome positioning inaccuracies, a mixed-sensitivity robust controller ($H_\infty$) is implemented, yielding a 33\% improvement in settling time and a 60\% reduction in steady-state error compared to classical PI control. Briefly explaining, $H_\infty$ control is a mathematical framework used in control theory to design robust systems that maintain stability and guaranteed performance even in the presence of uncertainties, unexpected disturbances, or against incomplete modelings. It works by framing the design process as an optimization problem—specifically, minimizing the maximum impact (the $H_\infty$ norm) that a worst-case disturbance or modeling error can have on the system's outputs \cite{Feng2023}. Finally, an adaptive Laplacian variance algorithm drives the intelligent autofocus, achieving optimal focal planes within 2 seconds. The operational block diagram in Fig.~\ref{setup} delineates the hardware architecture and software integration of the automated high-throughput image acquisition system.

\subsection{Object detection via \texorpdfstring{$U^{2}$-Net}{U2-Net} }

Pollen grain localization is performed on-the-fly using $U^{2}$-Net, a nested U-structure architecture optimized for salient object detection that generates pixel-level salience maps without the need for bounding box annotations \cite{qin2020u2net}. By applying morphological filtering to the saliency maps (area, circularity, and aspect ratio), the system reliably isolates target objects.

\section{Deep Metric Learning Classification}
\subsection{Architecture and Preprocessing}
\begin{figure}[htbp]
    \centering
\includegraphics[width=0.45\textwidth]{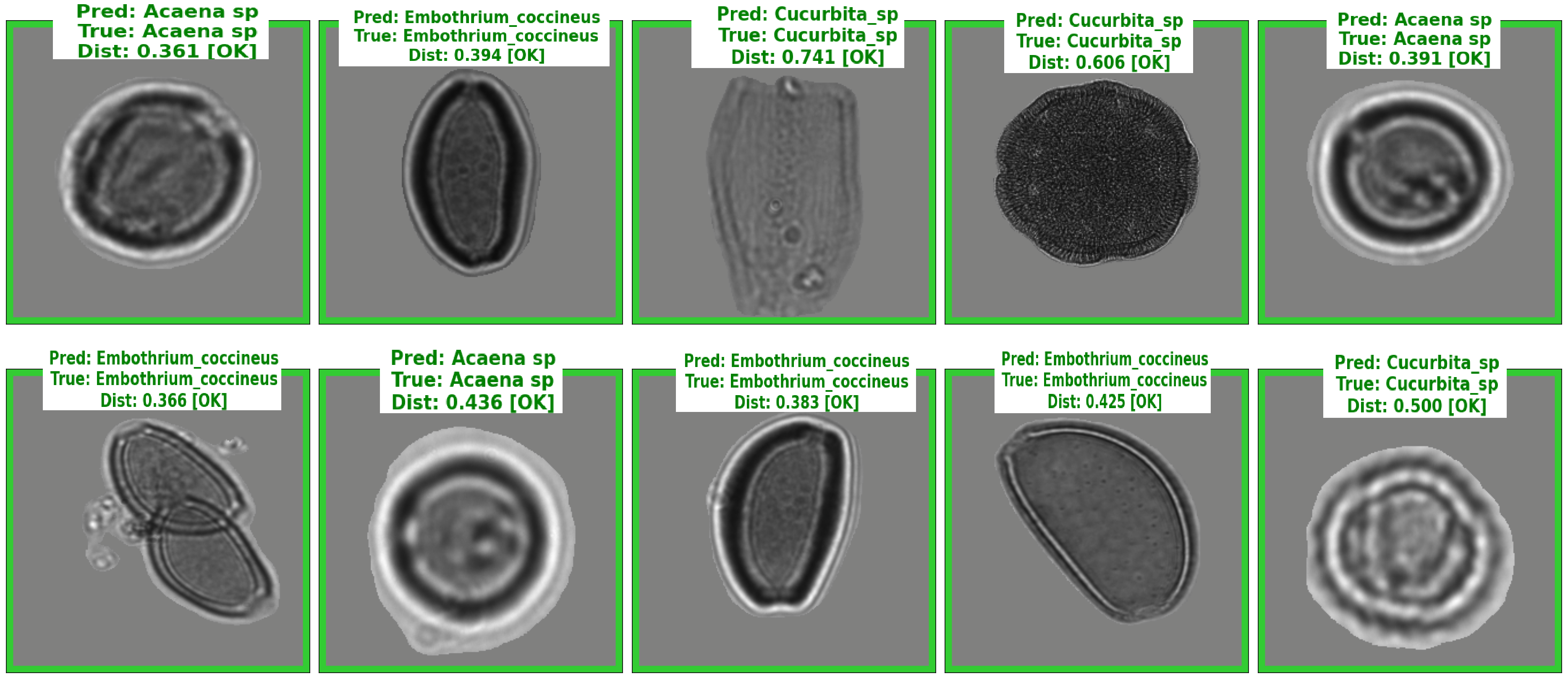}
    \caption{Predicted versus true: Inference preview examples - 10/10 correct. Best model (DinoV2) centroid-based classification. By normalizing the gray background (128), the resulting pixel values are almost zero. This means that the background does not trigger the neural network, eliminating the need for computationally expensive masked pooling.}
    \label{inference}
\end{figure}

Before classification, individual pollen grains are isolated using a $U^2$-Net detection pipeline. The network processes 320x320 monochrome images to generate a salient object probability map $P(x,y) \in [0,1]$. This map is binarized using an adaptive threshold $T = \mu(P) + k \cdot \sigma(P)$, and refined via 3x3 morphological operations. Extracted contours are filtered by area (5000–120,000 px) and circularity ($C \ge 0.3$). Localized crops are then classified using a DINOv2 ViT-S/14 backbone. Crucially, a normalized Slate Gray background (monochrome 128) acts as a zero-activation mask, bypassing computationally expensive masked pooling (Fig.~\ref{inference}).

\subsection{End-to-End Inference Pipeline}

\begin{figure*}[htbp]
    \centering
\includegraphics[width=0.9\textwidth]{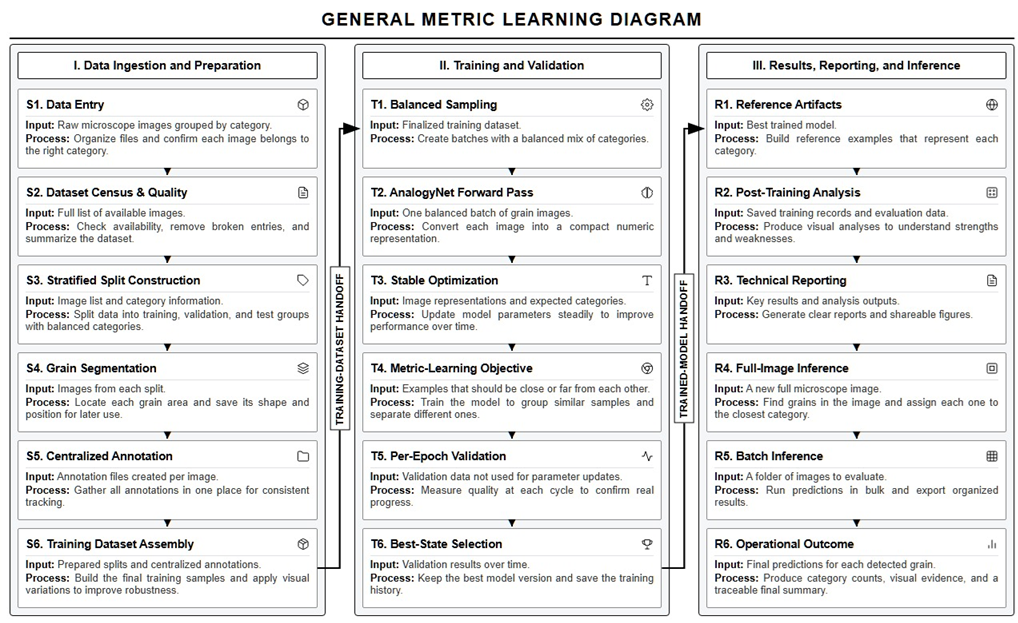}
    \caption{End-to-end schematic of the Metric Learning lifecycle, illustrating the sequential stages from raw microscope image ingestion through stable model optimization to final operational reporting and inference. The process is divided into three core phases: I. Data Ingestion and Preparation, which covers dataset curation and segmentation; II. Training and Validation, focusing on balanced sampling and model optimization using AnalogyNet; and III. Results, Reporting, and Inference, detailing the transition from post-training analysis to operational deployment and batch inference.}
    \label{Pipeline}
\end{figure*}

To seamlessly integrate detection and classification into a unified operational workflow, we implemented the Full Image Classifier pipeline (see schemes 1-5 in  supplementary material for further details). A general learning diagram is shown in Fig.~\ref{Pipeline}, which processes raw camera images of arbitrary resolutions through to final species counting. The inference process begins with the PollenGrainDetector system function (powered by the $U^{2}$-Net architecture), which localizes $N$ individual pollen grains within the full field of view, extracting their respective bounding boxes, geometric contours, and saliency scores. Subsequently, the pipeline iterates over each detected grain to extract a localized crop with a 20\% padding margin. The previously defined contour mask is applied to impose the standardized Slate Gray background. These standardized crops then undergo a deterministic validation transformation—consisting of a resize to 252x252 pixels and ImageNet normalization—before being fed into the AnalogyNet backbone. The network maps the input to a 128-dimensional unit hypersphere embedding, $\hat{e} \in \mathbb{S}^{127}$. 

Classification is executed using a nearest-centroid approach, where the predicted class $\hat{y}$ is determined by minimizing the L2 distance between the grain's embedding and the pre-computed class centroids. Alongside the predicted class, the system calculates the absolute distance and a confidence score. Ultimately, the pipeline aggregates these per-grain predictions to generate three distinct outputs: a fully annotated microscopy image featuring class-color-coded bounding boxes and labels, a quantitative palynological count mapping each species to its respective frequency, and optional per-grain Gradient-Weighted Attention heatmaps for enhanced interpretability.

\subsection{Image annotation, segmentation, and background}
To maximize system efficiency, the architecture implements a ``zero-additional-image" design principle. Instead of generating and storing physical image crops, the pipeline utilizes plain-text annotation files containing metadata and per-grain parameters, such as bounding boxes, saliency scores, and contour coordinates. Within this workflow, a custom dataset class dynamically extracts pollen crops during the training and inference phases to eliminate redundant storage, while a 64-image Least Recently Used (LRU) cache optimizes access speeds by managing original high-resolution imagery in memory. Furthermore, each detected grain is extracted with a 20\% padding margin, resized to 252x252 pixels, and subjected to contour masking.The background suppression strategy involves replacing the masked area with a Slate Gray value (128), which, following standard ImageNet normalization, functions mathematically as a null vector:$$\text{gray}_{norm} = \frac{128/255 - \mu}{\sigma} \approx (0.07, 0.02, -0.03) \approx \mathbf{0}$$By approximating a zero-vector, the background generates near-zero activations within the neural network backbone. This approach effectively achieves visual noise suppression and minimizes background contribution to global pooling operations without the computational overhead associated with explicit masked pooling techniques.

\subsection{Preprocessing and Augmentations}

\begin{figure}[htbp]
    \centering
\includegraphics[width=0.45\textwidth]{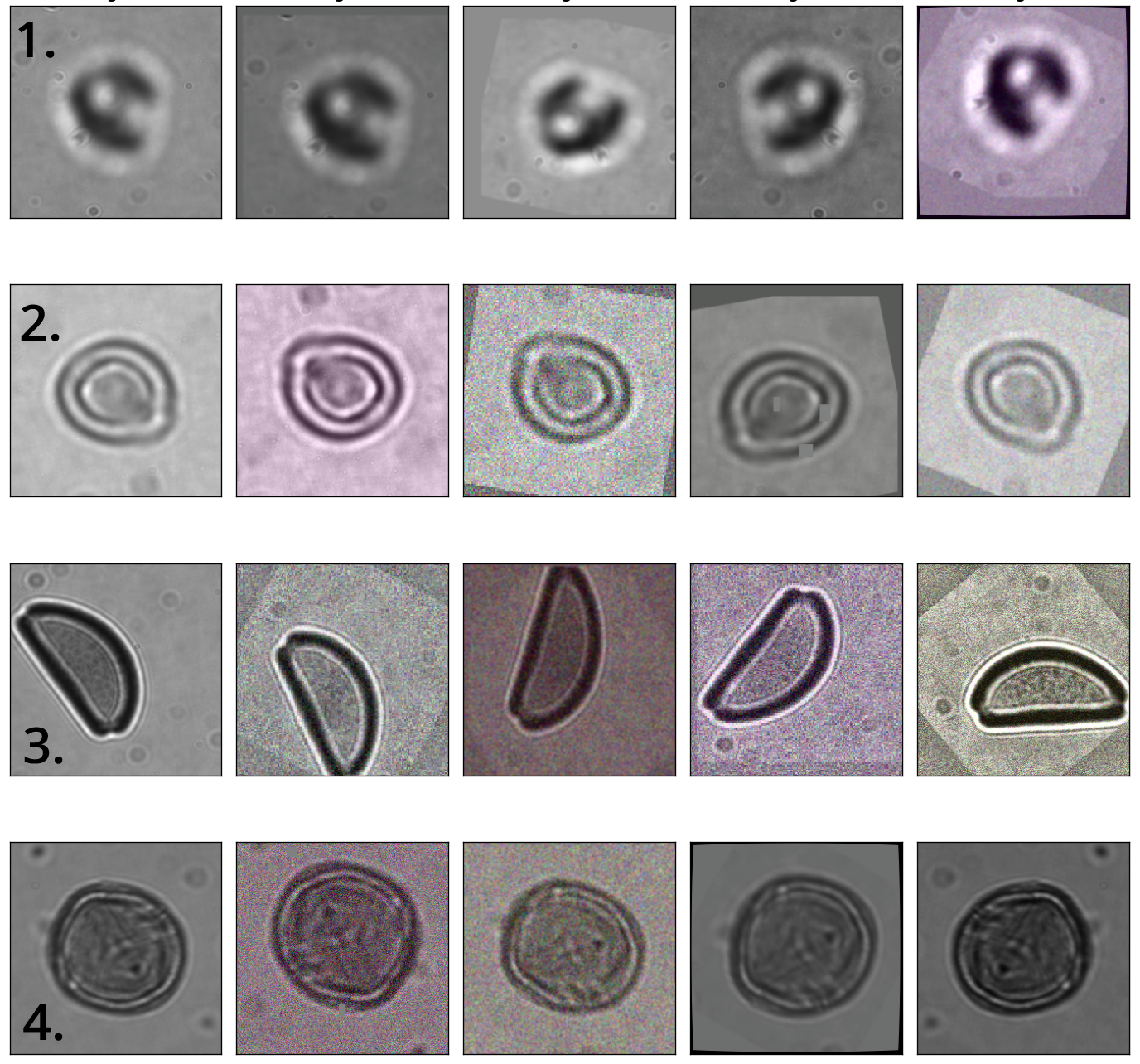}
    \caption{Examples of four synthetic augmentation performed to left starting from the originals 1. \textit{Acaena splendens} (bidibid), 2. \textit{Cucurbita pepo} (squash), 3. \textit{Embothrium coccineum} (Chilean firebush) and 4. \textit{Lithrea caustica }(litre tree) }
    \label{augmentation}
\end{figure}

To enhance robustness against microscopy variations, the training pipeline employs a comprehensive augmentation strategy via the Albumentations library \cite{Shorten2019, Buslaev2020,Alomar2023,Kumar2025} (Fig.~\ref{augmentation}). Geometric transformations—including random resized crops, flips, rotations up to $180^\circ$, and affine translations—account for magnification shifts and isotropic orientation. Optical and sensor artifacts are simulated through color jittering, CLAHE, Gaussian noise, and blurring to model inconsistent staining and hardware noise. CoarseDropout is further utilized to mimic physical occlusions. Critically, to preserve the zero-activation background, all borders and dropout regions are filled with a uniform gray value (monochrome: 128). Post-augmentation, images are normalized using ImageNet parameters. In contrast, the validation pipeline remains deterministic, utilizing only resizing and normalization to ensure objective performance assessment.

\subsection{Embedding Model }

To construct a robust embedding space, AnalogyNet utilizes a DINOv2 ViT-S/14 backbone, which offers a global receptive field and universal feature extraction via self-supervised pre-training. Despite its smaller footprint (21M parameters vs. ResNet-50’s 25.5M), this Transformer-based architecture provides superior efficiency and emergent segmentation capabilities. The pipeline (Fig. 4 in Supp. material) processes a $252 \times 252$ monochrome image into an $18 \times 18$ grid of 324 patch tokens (plus one CLS token) with a hidden dimension of 384. These propagate through 12 encoder blocks featuring multi-head self-attention (6 heads) and SwiGLU activations. To derive the final representation, the network applies mean pooling over the spatial patch tokens, yielding a 384-dimensional vector. This vector is passed through a three-layer MLP Projection Head to reach a 128-dimensional latent space. Finally, $L_2$ normalization projects the embedding onto a unit hypersphere ($S^{127}$), ensuring that the squared Euclidean distance is monotonically equivalent to cosine similarity:$$\|a - b\|_2^2 = 2(1 - \cos(a, b))$$This geometric constraint allows the model to optimize taxonomic similarities with high precision and consistency.

\subsection{Multi-Similarity Loss}
To optimize the mapping of pollen crops into the $\mathbb{S}^{127}$ embedding space, we train the DINOv2 backbone using the Multi-Similarity (MS) Loss. Traditional metric learning objectives, such as Triplet Loss, suffer from slow convergence and sparse gradient updates because they only connect an anchor to a single positive and a single negative sample at a time. In contrast, the MS Loss computes dense gradients by simultaneously connecting the anchor point to all positive samples (pulling them together) and all negative samples (pushing them apart) within a given mini-batch. Mathematically, the Multi-Similarity Loss is formulated in \cite{wang2019multi} as:

\begin{align}
\mathcal{L}_{MS} &= \frac{1}{m} \sum_{i=1}^{m} \left\{ \frac{1}{\alpha} \log \left[ 1 + \sum_{k \in \mathcal{P}_i} e^{-\alpha(S_{ik} - \lambda)} \right] \right. \\
&+ \frac{1}{\beta} \log\left[ 1 + \sum_{k \in \mathcal{N}_i} e^{\beta(S_{ik} - \lambda)} \right] \nonumber
\end{align}

where $m$ denotes the number of anchors in the batch, while $\mathcal{P}_{i}$ and $\mathcal{N}_{i}$ represent the sets of positive and negative instances for the $i$-th anchor, respectively. The term $S_{ik}$ corresponds to the cosine similarity between the embeddings of the anchor and the $k$-th sample.This formulation utilizes a margin shift parameter, $\lambda = 0.5$, to define the similarity threshold for mining informative pairs. Additionally, the scaling factors $\alpha = 2.0$ and $\beta = 50.0$ control the exponential weighting of the similarities, allowing the network to dynamically assign higher gradient weights to the hardest positive and negative pairs during backpropagation. By leveraging this dense pulling and pushing mechanism, the model efficiently constructs a highly separable geometric latent space for taxonomic classification.

\section{Evaluation and Metrics}

\begin{figure}[htbp]
    \centering
\includegraphics[width=0.4\textwidth]{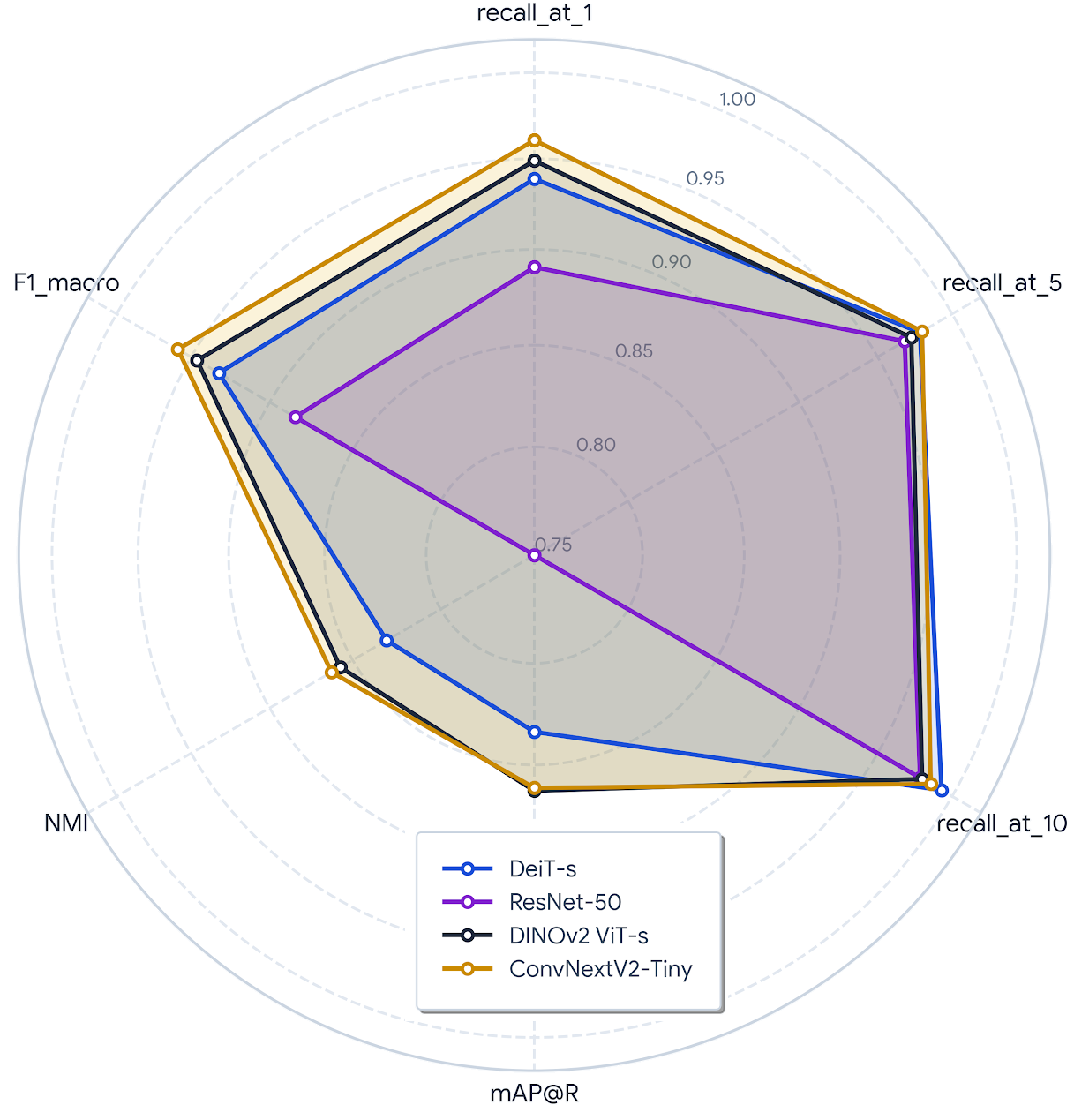}
    \caption{Performance comparison. Several baseline models were considered, including DeiT-s, ResNet-50, DINOV2 VIT-s and ConvNextV2-tiny algorithms. These models provide a representative range of commonly used approaches. Final training indicators are used to compare the models. A logarithmic scale is used for the intervals to be compared, with a minimum of 0.75 to 1.}\label{spider}
\end{figure}

\begin{figure}[htbp]
    \centering
    \includegraphics[width=\columnwidth]{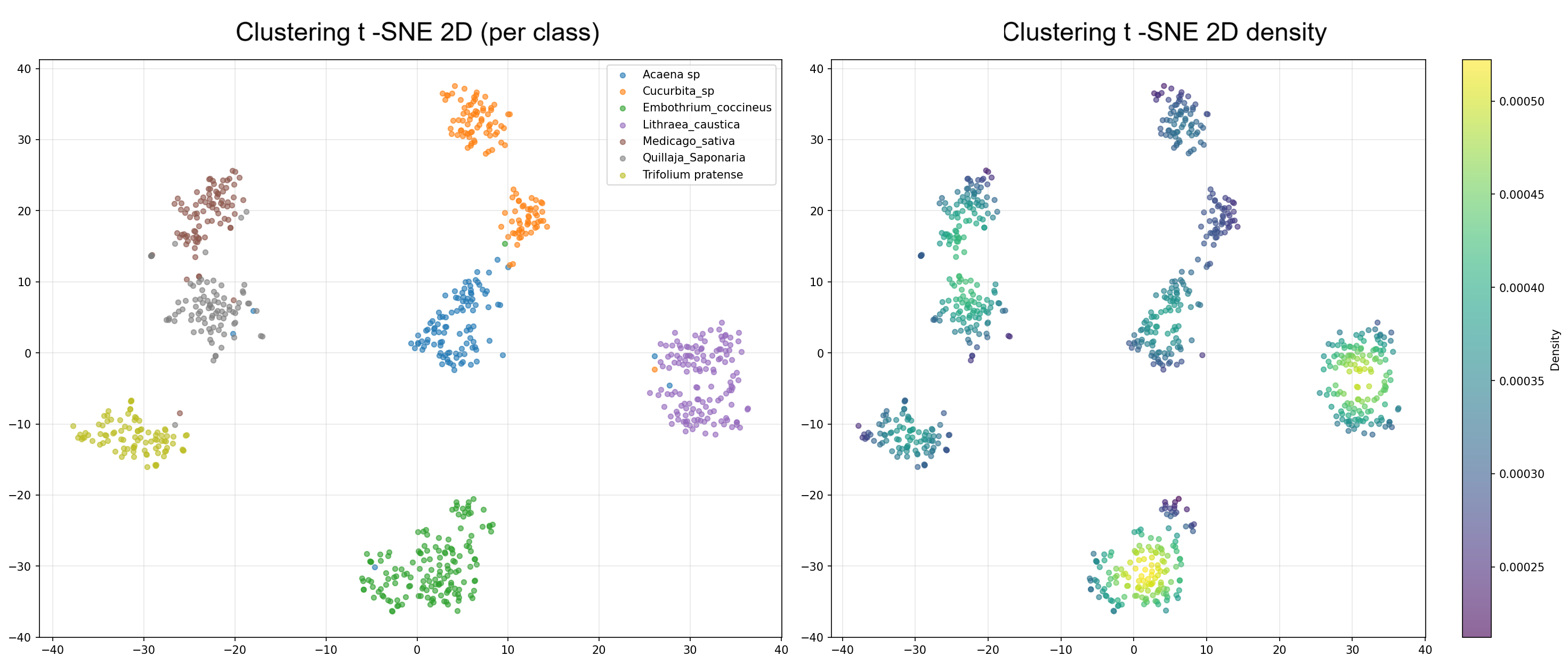}
    \caption{Model comparison based on Latent Space Analysis: 2D t-SNE per class and density of 685 samples. Blocks apply for DINOv2.}
    \label{t-sne2D}
\end{figure}

\begin{figure}[htbp]
    \centering
\includegraphics[width=0.45\textwidth]{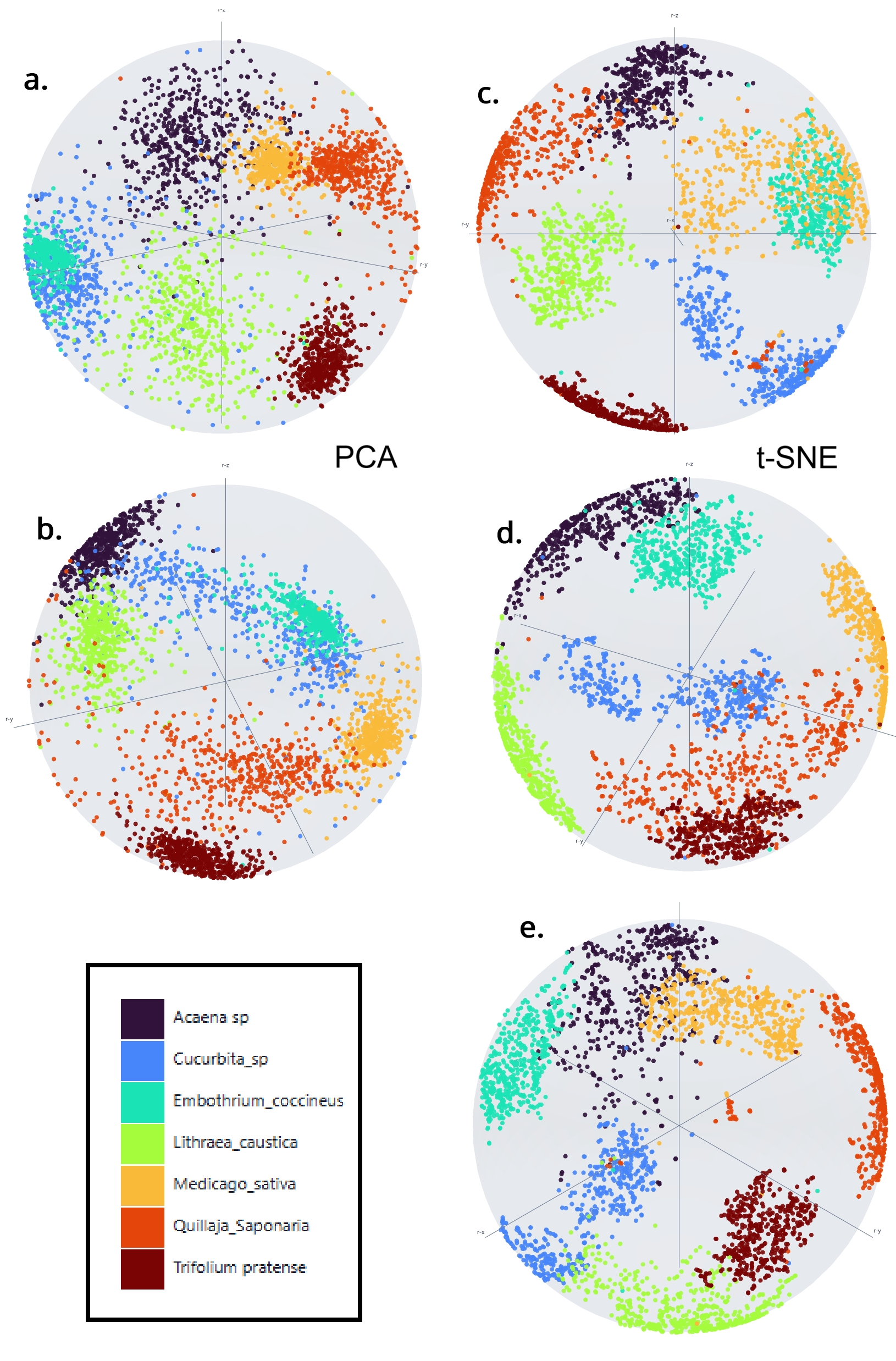}
    \caption{Hypersphere embedding representation in DINOv2 ViT-s. (a) and (b).: Two 3D PCA views. (c), (d) and (e) Three 3D t-SNE angular perspectives. Block: Color code by class. Clear geometric separability between both algorithms (PCA vs t-SNE). This Figure explicitly illustrates the geometric constraint imposed at the end of the DINOv2 model pipeline showing clusters projections onto the surface of a unit hypersphere.}
    \label{Hypershere_embedding3D}
\end{figure}

\begin{figure*}[!t]
\centering
\subfloat[]{\includegraphics[width=3in]{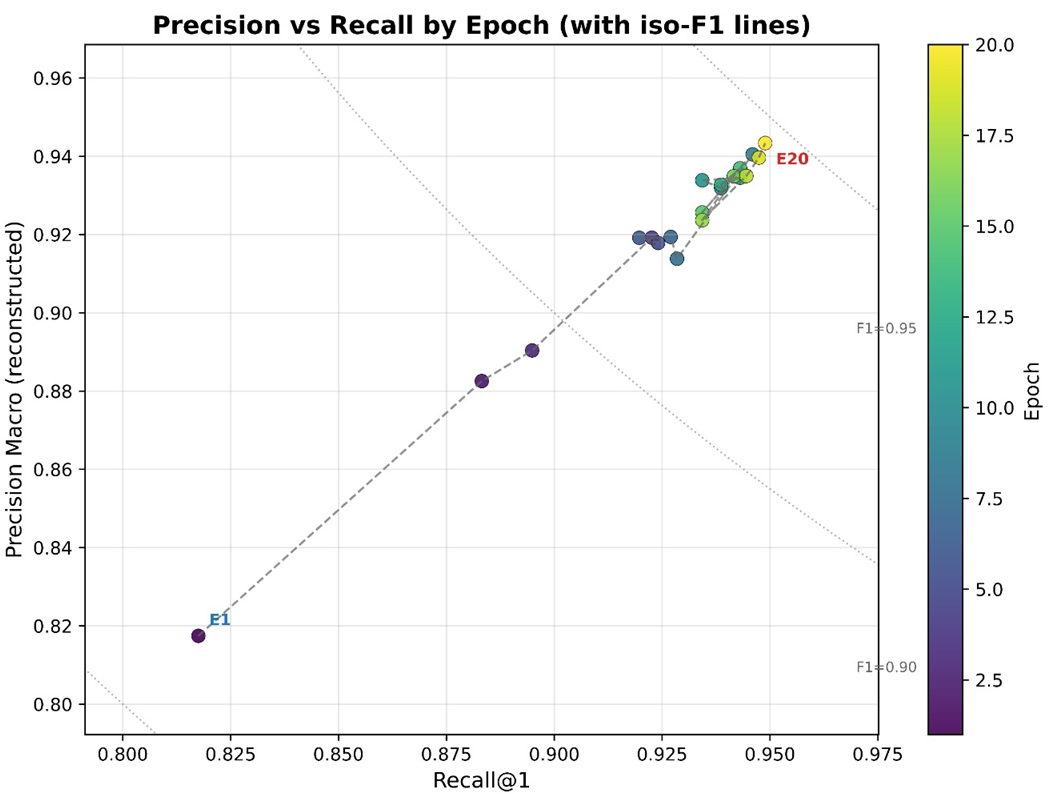}%
\label{precision_recall_F1_curves}}
\hfil
\subfloat[]{\includegraphics[width=4 in]{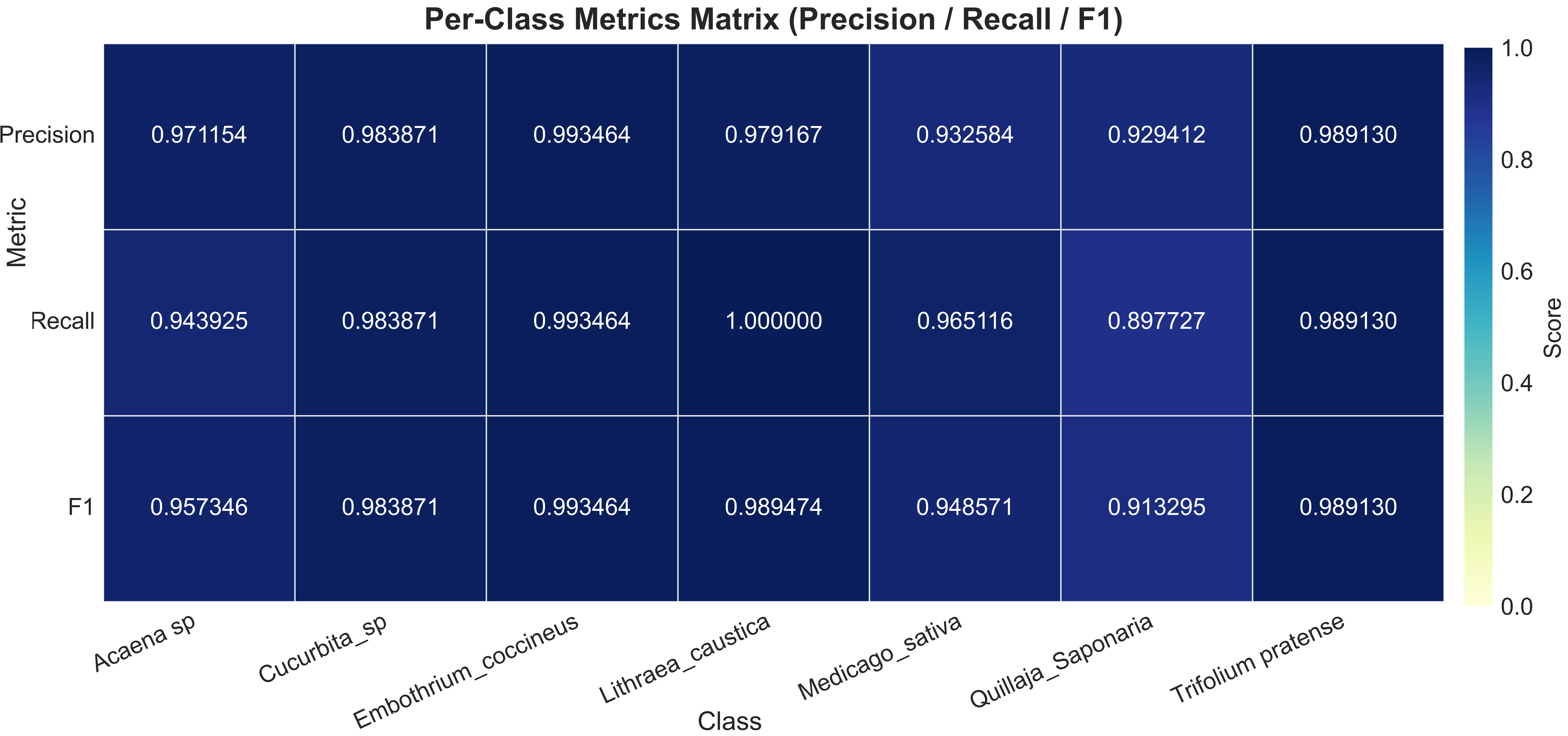}%
\label{precision_recall_F1_metric}}
\caption{Quantitative results in DINOv2 ViT-s: Performance metrics. (a) Precision versus recall with iso-Fi1 curves by epoch. (b) Per class metrics matrix; Precision, recall and F1 score.}
\label{fig_sim}
\end{figure*}

To stablish a comparative reference, four models were used to evaluates different backbone architectures including ResNet-50, ConvNeXt V2-tiny, DeiT-s, and DINOv2 ViT-s. DINOv2 ViT-S/14 is the optimal and most computationally efficient configuration (21M parameters vs. ResNet-50's 25.5M) (Fig.~\ref{spider}, Table~\ref{tab:table1}). Pretrained via Self-Supervised Learning (SSL) on 142M images, DINOv2 establishes a global receptive field immediately and exhibits emergent segmentation that outperforms baselines across latent space metrics like Recall@K, NMI, and mAP@R.  By applying $L_2$ normalization ($\hat{e} = e/||e||_2$), features are strictly projected onto a 128-dimensional hypersphere ($\mathbb{S}^{127}$), mathematically linking squared Euclidean distance to cosine similarity for consistent taxonomic classification (Figs.~\ref{t-sne2D} and \ref{Hypershere_embedding3D}). Performance is rigorously assessed using metrics such as Accuracy, Precision, F1-score, and Recall@K (the fraction of query images returning a correct match within the top-K results, Fig.~\ref{precision_recall_F1_metric}).

    \begin{table*}[htbp] 
    \centering
    \caption{Backbone comparison (global)\label{tab:table1}}
    \begin{tabular}{rlccccccc}
        \toprule
        \textbf{Rank} & \textbf{Model} & \textbf{Accuracy} & \textbf{B. Acc} & \textbf{Prec. M} & \textbf{Prec. W} & \textbf{F1 M} & \textbf{F1 W} & \textbf{Dist. Ratio} \\
        \midrule
        1 & ConvNextV2-Tiny & 0.9747 & 0.9706 & 0.9718 & 0.9748 & 0.9711 & 0.9746 & 1.6786 \\
        2 & DINOv2 ViT-s & 0.9722 & 0.9676 & 0.9684 & 0.9721 & 0.9679 & 0.9721 & 1.8418 \\
        3 & DeiT-s & 0.9633 & 0.9565 & 0.9579 & 0.9639 & 0.9566 & 0.9632 & 1.5621 \\
        4 & ResNet-50 & 0.9064 & 0.8996 & 0.9008 & 0.9060 & 0.8998 & 0.9059 & 1.2683 \\
        \bottomrule
    \end{tabular}
\end{table*}

By limiting the data to spherical surfaces (where the magnitude of all vectors is 1), the squared Euclidean distance becomes mathematically linked to cosine similarity. Therefore, Fig.~\ref{Hypershere_embedding3D} visualizes how these classes are mathematically structured to ensure that the distance calculation is geometrically consistent and accurate for taxonomic classification. This Figure represents the step in which an L2 normalization ($\hat{e}=e/||e||_{2}$) is applied to the extracted features, which forces all embeddings to project strictly onto the surface of a 128-dimensional hypersphere, denoted as $\mathbb{S}^{127}$.

To rigorously assess the performance of the proposed Deep Metric Learning model  DINOv2 ViT-S/14, a comprehensive suite of evaluation metrics is employed.  The primary retrieval metric is Recall@K (Fig.~\ref{precision_recall_F1_metric}), defined as the fraction of query images that successfully return at least one correctly matched neighbor within the top-K retrieved samples. Mathematically, Recall, Accuracy, Precision, and F1- scores are defined as follows

\begin{align}
\vn{Accuracy}  &= \frac{TP+TN}{TP+TN+FP+FN}\\[\jot]
\vn{Precision} &= \frac{TP}{TP+FP}\\[\jot]
\vn{Recall}    &= \frac{TP}{TP+FN}\\[\jot]
F_1            &= \frac{2\cdot\vn{Precision}\cdot\vn{Recall}}%
                       {\vn{Precision}+\vn{Recall}}
\end{align}

\begin{figure}[htbp]
    \centering
\includegraphics[width=0.45\textwidth]{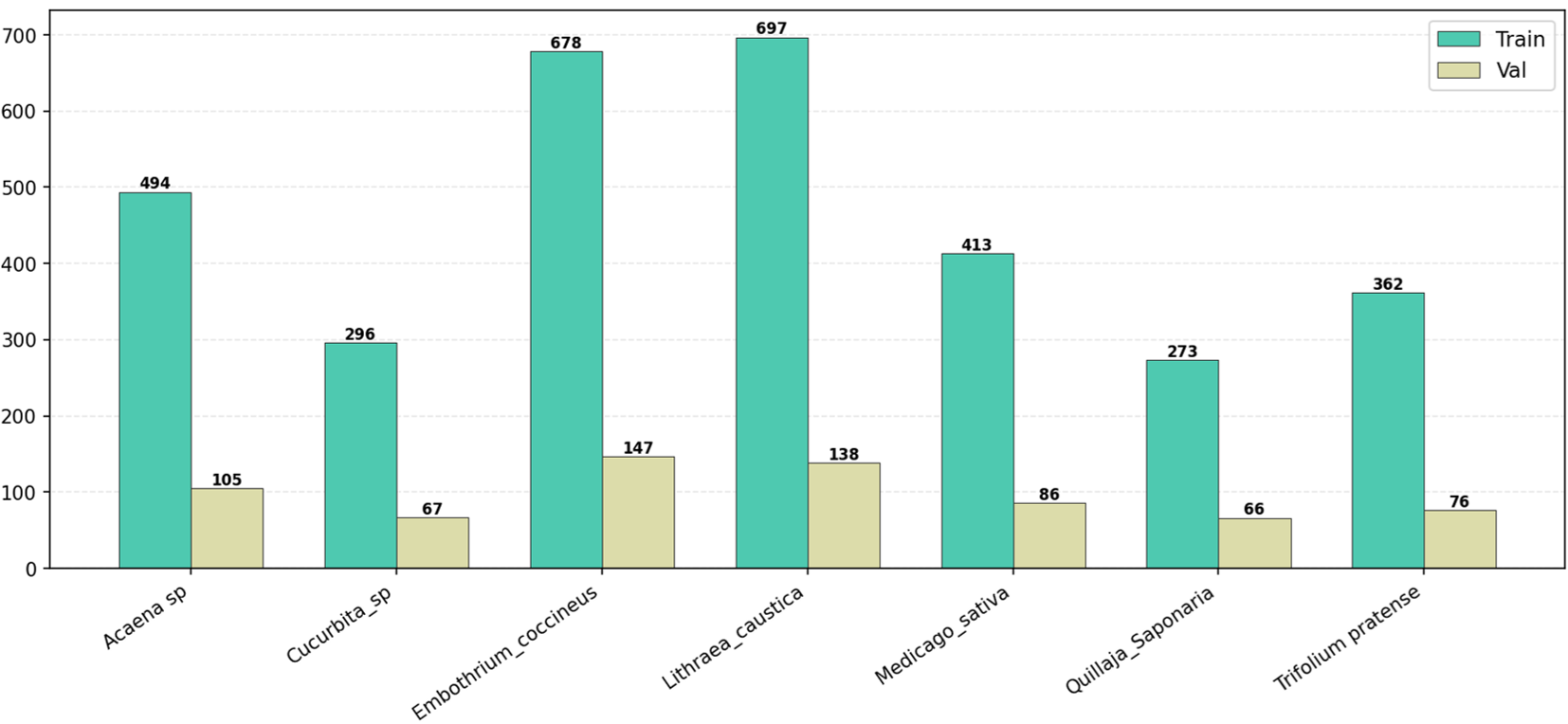}
    \caption{Train and validation balance in DINOv2 ViT-s model.}
    \label{trainValance}
\end{figure}

\begin{figure}[htbp]
    \centering
\includegraphics[width=0.40\textwidth]{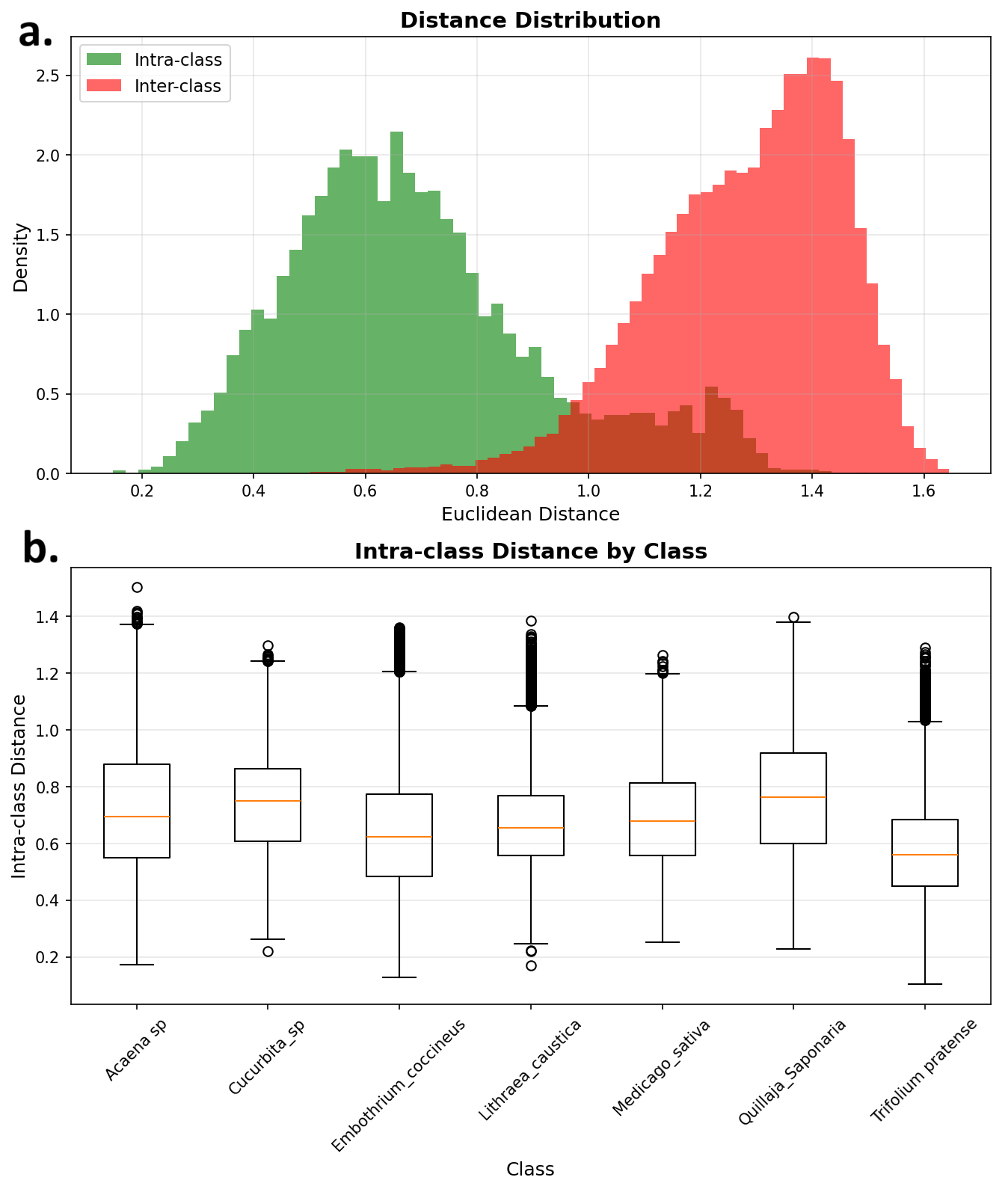}
    \caption{a. Distance distribution in DINOv2 ViT-s.  Intra and inter-class distance. b. Intra-class distance by class.}
    \label{distribution2}
\end{figure}

\begin{figure}[htbp]
    \centering
\includegraphics[width=0.5\textwidth]{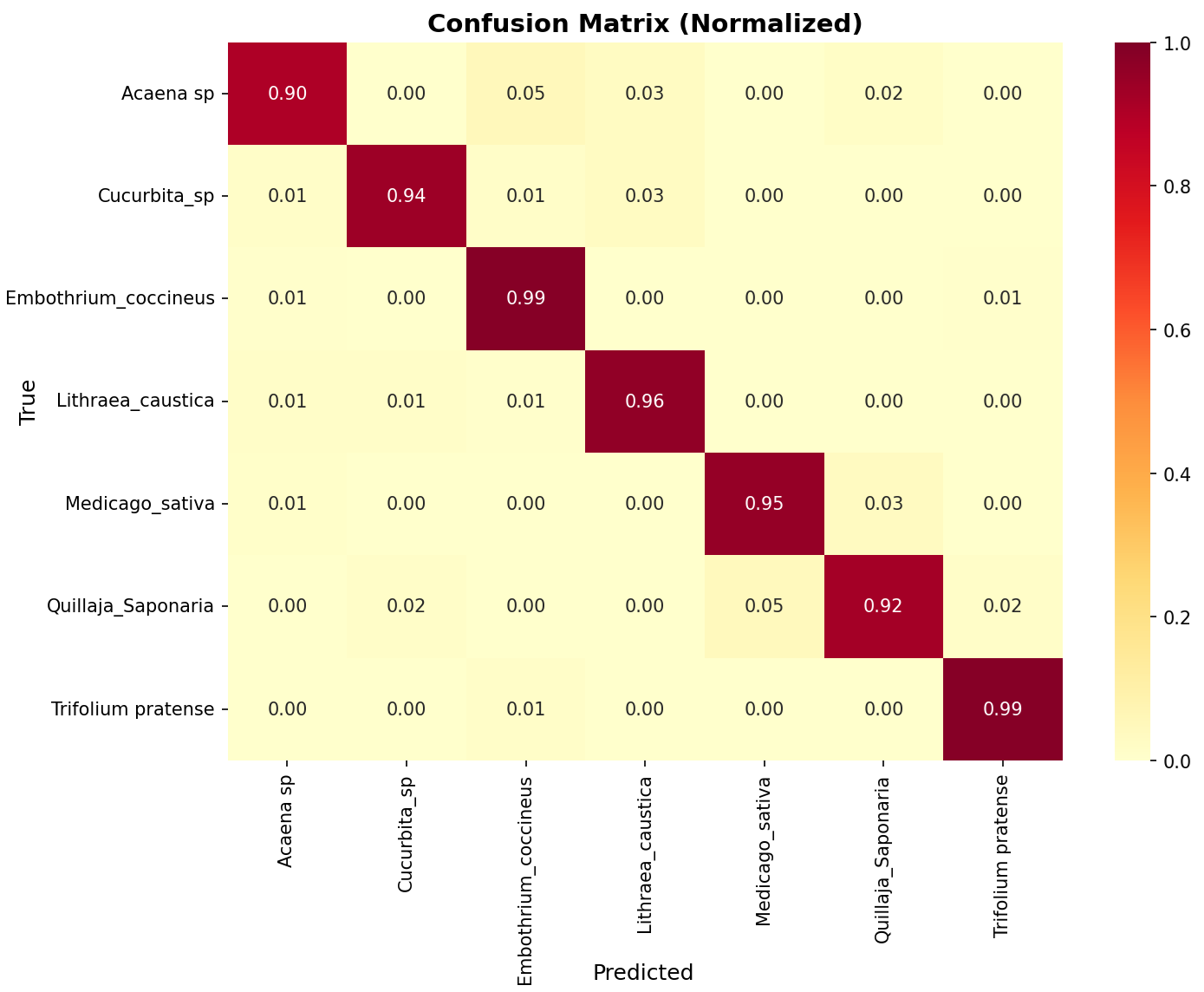}
    \caption{a. b. Normalized confusion matrix for DINOv2 ViT-s from 0 to 1. 1 is 100\% accuracy (95.1\% Accuracy global).}
    \label{cmatrix}
\end{figure}

Using standard TP/TN/FP/FN values, latent space integrity is evaluated via Normalized Mutual Information (NMI) and a Distance Ratio ($\rho = \bar{d}_{inter} / \bar{d}_{intra}$), targeting $\rho > 3$ for robust separability. Training balance, distance distributions, and kNN validation metrics are detailed across Figs. Fig.~\ref{Hypershere_embedding3D}, ~\ref{trainValance}-~\ref{cmatrix}, and Tables \ref{tab:table2}-\ref{tab:table3}. The system's final predictive assignment utilizes Centroid Classification, mapping query samples to the closest learned centroid by minimizing the Euclidean distance.

\begin{table}[!t]
\caption{Standard Classification Metrics\label{tab:table2}}
\centering
\begin{tabular}{|c||c|}
\hline
Metric & Value\\
\hline
Accuracy & 0.9722 (97.2\%)\\
\hline
Balanced Accuracy & 0.9676 (96.8\%)\\
\hline
Precision Macro & 0.9684\\
\hline
Precision Weighted & 0.9721\\
\hline
F1 Macro & 0.9679\\
\hline
F1 Weighted & 0.9721\\
\hline
\end{tabular}
\end{table}

\begin{table}[!t]
\caption{Per-Class Accuracy (KNN)\label{tab:table3}}
\centering
\begin{tabular}{|c||c|}
\hline
Metric & Value\\
\hline
\textit{Acaena splendens} & 94.4\%\\
\hline 
\textit{Cucurbita pepo} & 98.4\% \\
\hline
\textit{Embothrium coccineum} & 99.3\% \\
\hline
\textit{Lithrea caustica } & 100.0\% \\
\hline
\textit{ Medicago sativa} & 96.5\% \\
\hline
\textit{Quillaja saponaria} & 89.8\%\\
\hline
\textit{Trifolium pratense} & 98.9\%\\
\hline
\end{tabular}
\end{table}

Classification accuracy is further validated using a k-Nearest Neighbors (kNN) Confusion Matrix shown in Fig.~\ref{cmatrix}, constructed via a leave-one-out evaluation strategy over the validation embeddings. 

Finally, the ultimate predictive assignment of the system is formalized through Centroid Classification. In this operational phase, a query sample $x$ mapped through the embedding function $f_\theta$ is assigned to the class $\hat{y}$ of the closest learned centroid $\hat{\mu}_c$ by minimizing the Euclidean distance:

\begin{equation}
\hat{y} = \text{argmin}_c \|f_\theta(x) - \mu_c\|_2
\end{equation}

\subsection{Interpretability and Gradient-Weighted Attention (XAI)}

To provide transparency, the system extracts high-fidelity explainability heatmaps directly from the DINOv2 backbone. Unlike traditional Grad-CAM, which often yields diffuse results for Vision Transformers, our approach leverages the inherent self-attention mechanisms of the final transformer block.Using PyTorch hooks, we extract the CLS-to-Patch attention matrix, $A = \text{softmax}(QK^T / \sqrt{d_k})$. To ensure these visualizations focus on task-specific morphological features, we implement a Gradient-Weighted Attention mechanism. We compute the importance weight $h_j$ for each attention head by tracking the gradient of the cosine similarity between the grain embedding and its predicted class centroid with respect to the $QKV$ activations:$$h_j = \text{mean} \left( \left| \frac{\partial \text{sim}}{\partial QKV_j} \right| \right)$$The final heatmap is a weighted sum of the head-specific attentions, $A_{\text{weighted}} = \sum_{j} (h_j \cdot A_{cls,j})$, upsampled to $252 \times 252$. This process precisely isolates diagnostic features—such as rugose textures or germinal pores—making the model’s internal logic visually accessible to human experts.

\begin{figure}[htbp]
    \centering
\includegraphics[width=0.45\textwidth]{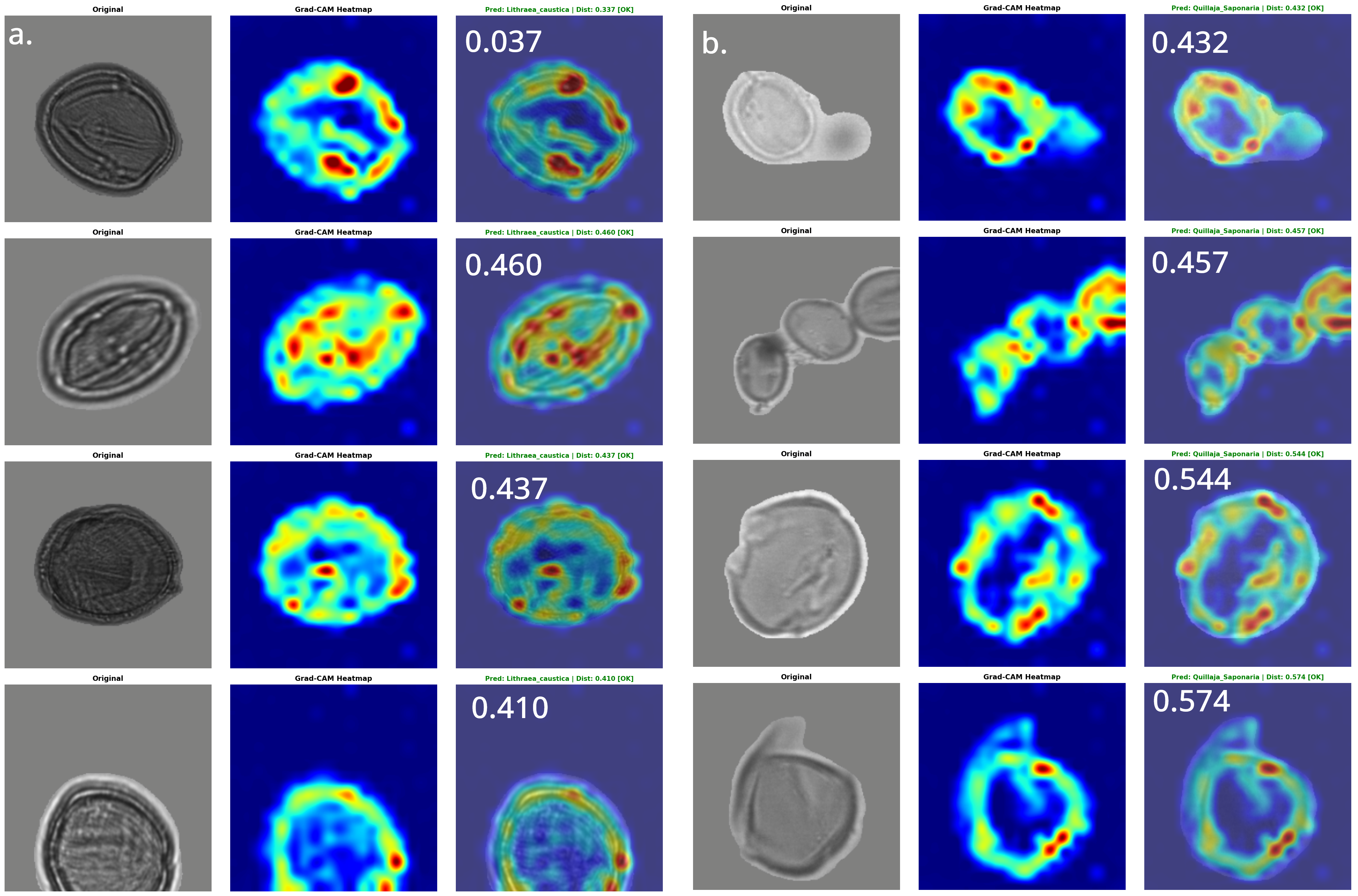}
    \caption{DINOv2 ViT-s Grad-CAM heatmaps for endemic pollen: a. \textit{Lithrea caustica} and b. \textit{Quillaja saponaria}. Displays original images alongside predicted Euclidean distances in the $\mathbb{S}^{127}$ latent space. Through Centroid Classification on L2-normalized embeddings, the predicted labels correspond to the class centroid minimizing this distance.}
    \label{gradcam}
\end{figure}

\begin{figure}[htbp]
    \centering
\includegraphics[width=0.3\textwidth]{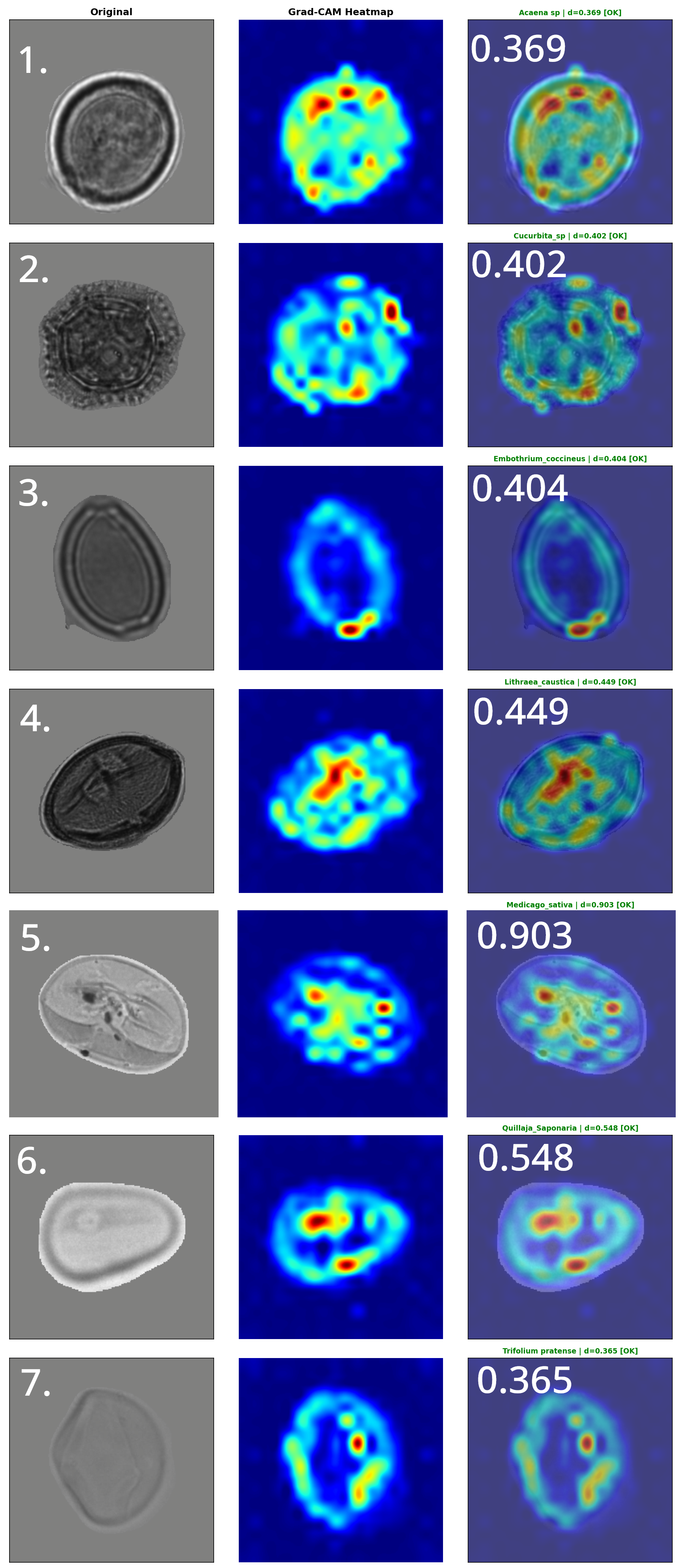}
    \caption{DINOv2 ViT-s Grad-CAM summary for all 7 classes (1. \textit{A. splendens}, 2. \textit{C. pep}o, 3. \textit{E. coccineum}, 4. \textit{L. caustica}, 5. \textit{M. sativa}, 6. \textit{Q. saponaria}, 7. \textit{T. pratense}). Heatmaps highlight diagnostic textures, ornamentation, and morphological shapes. Numerical values display the predicted Euclidean distance within the $\mathbb{S}^{127}$ latent space. }
    \label{gradcam2}
\end{figure}

Fig.~\ref{gradcam} are Grad-CAM examples for \textit{Lithrea caustica} and \textit{Quillaja saponaria}, the numerical values for prediction refer to the final classification and geometric validation steps of the Deep Metric Learning pipeline. The Euclidean distance is the distance between the embedding vector of the specific pollen grain being analyzed and the pre-computed centroid of its assigned taxonomic class. A smaller distance indicates that the morphological features of the grain (captured in the Grad-CAM heatmap) are very similar to the ``ideal" prototype of that species. For example, the value $0.037$ indicates an extremely high similarity to the \textit{Lithrea caustica} centroid. Gradient-Weighted Attention summary for all classes is exhibited in Fig.~\ref{gradcam2}.

To obtain morphological values, the effective spatial resolution (pixel size ($Px_S$) in object space equal to resolution object $Rs_O$) is determined by the ratio between the physical pixel size of the CMOS sensor and the total magnification of the optical system ($M_{OS}$). For a system utilizing a 60x oil immersion objective and a scientific CMOS sensor with a physical pixel pitch of $3.774~\mu$m, the projected resolution is calculated as:$$Rs_O=\frac{Px_S}{M_{OS}}$$ Given these parameters, the resolution is exactly $0.0629~\mu$m/px, as empirically verified using a USAF 1951 Test Target from Thorlabs. This spatial sampling ensures compliance with the Nyquist-Shannon criterion for resolving the diffraction-limited features of the pollen grains, such as exine ornamentation and germinal pores identified through Gradient-Weighted Attention.

\begin{table*}[t]
\centering
\caption{Morphological and Texture Metrics for Pollen Identification}
\label{tab:pollen_stats_short}
\begin{tabular*}{\textwidth}{@{\extracolsep{\fill}}lcccc}
\toprule
\textbf{Pollen Class} & \textbf{Size} ($\mu$m) & \textbf{Area} ($10^3$ px$^2$) & \textbf{Circ.} ($C$) & \textbf{I.C.D.} ($\mu \pm \sigma$) \\
\midrule
1. \textit{Acaena s.} & $12.67 \pm 1.30$ & $45.2 \pm 5.4$ & $0.88 \pm .04$ & $0.68 \pm .12$ \\
2. \textit{Cucurbita p.} & $45.2 \pm 5.4$ & $108.5 \pm 12.1$ & $0.92 \pm .03$ & $0.74 \pm .10$ \\
3. \textit{Embothrium c.} & $20.17 \pm 1.89$ & $35.4 \pm 4.1$ & $0.65 \pm .06$ & $0.62 \pm .14$ \\
4. \textit{Lithrea c.} & $16.15 \pm 1.60$ & $41.0 \pm 4.8$ & $0.78 \pm .05$ & $0.65 \pm .11$ \\
5. \textit{Medicago s.} & $11.99 \pm 1.01$ & $28.1 \pm 3.2$ & $0.84 \pm .04$ & $0.67 \pm .13$ \\
6. \textit{Quillaja s.} & $14.77 \pm 1.39$ & $38.4 \pm 4.5$ & $0.86 \pm .04$ & $0.76 \pm .15$ \\
7. \textit{Trifolium p.} & $11.07 \pm 0.88$ & $22.3 \pm 2.8$ & $0.89 \pm .03$ & $0.58 \pm .11$ \\
\bottomrule
\multicolumn{5}{l}{\small \textit{Note:} I.C.D. refers to Intra-class Distance in the $\mathbb{S}^{127}$ latent space. Res: $0.15~\mu$m/px.}
\end{tabular*}
\end{table*}

\begin{figure}[htbp]
    \centering
\includegraphics[width=0.45\textwidth]{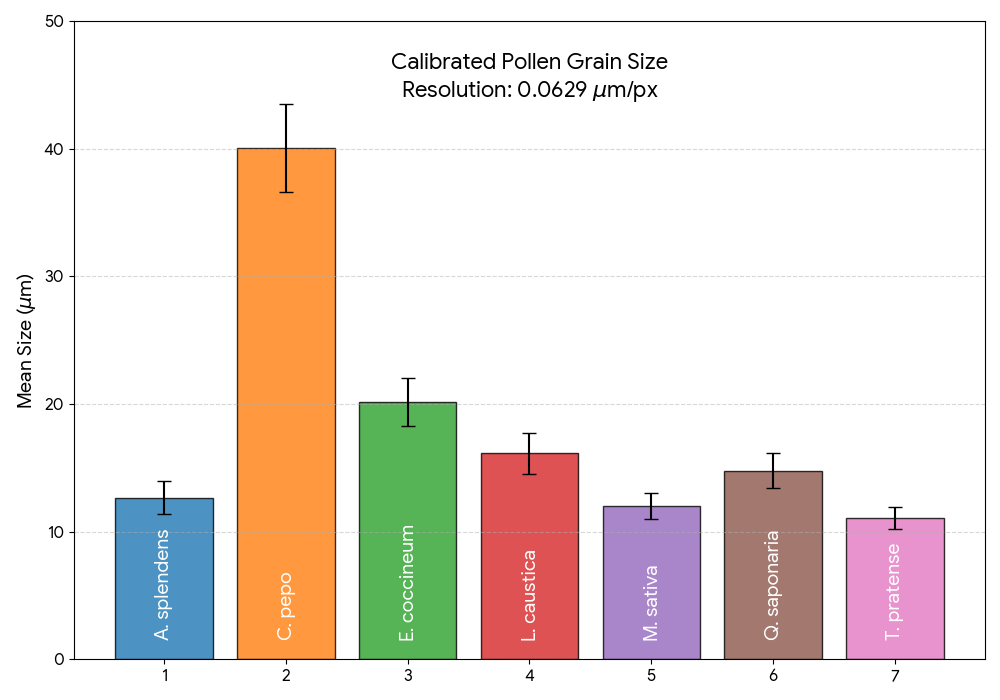}
    \caption{Mean morphological diameter ($\mu$) and standard deviation ($\sigma$) from $U^2$-Net masks (0.15~$\mu$m/px resolution, 60x objective). Ordered by size, the distribution illustrates the biological variability across endemic, native, and introduced flora, ranging from Cucurbita pepo ($45.2 \pm 5.4~\mu$m) to Trifolium pratense ($11.07 \pm 0.88~\mu$m).}
    \label{size}
\end{figure}

In Table~\ref{tab:pollen_stats_short} Size and Area values ($\mu \pm \sigma$) were calculated by applying the pixel-to-micron conversion to the area thresholds ($1,000$ to $120,000$ px$^2$) used by the $U^{2}$-Net detector. Circularity values ($C$) reflects the geometric robustness where $C = 4\pi A / P^2$. Classes like \textit{Cucurbita pepo} show high stability (low $\sigma$) due to their consistent spherical shape. Additionally, statistical distribution of pollen grain sizes across the seven analyzed classes from the Biobío region are shown in Fig.~\ref{size}

Important to mention that Intra-class Distance is a metric that represent the mean Euclidean distance of samples to their class centroid in the 128-dimensional embedding space. A lower standard deviation indicates higher taxonomic consistency in the model's feature extraction for that specific species. Roughness Interpretation is the standard deviation in circularity and intra-class distance often correlates with the rugose or irregular textures identified by the Grad-CAM heatmaps.

Concerning the operational efficiency, traditional melissopalynological analysis is a highly labor-intensive and operator-dependent task, where the manual identification of a single sample under a microscope can take from several hours to a full day, depending on pollen diversity and preservation \cite{Rostami2025}. Specifically, manual counts of merely 600 pollen grains can take between 2 and 4 hours for an expert palynologist \cite{Theuerkauf2026}. Given that the Chilean standard NCh2981.Of2005 requires a count of at least 1,200 grains to validate the botanical origin of honey \cite{García2020,Montenegro2008}, a human expert would need an estimated 4 to 8 hours per sample. In stark contrast, our automatic classifier, operating at an average of 126.0 ms per image, processes the 1,600 images required to reach the 1,200-grain target in approximately 3.36 minutes of pure computational time. Even when accounting for the mechanization of scanning displacement control, which may scale this overall time by a maximum of $\backsim$ 30 minutes per sample coverslip, the proposed solution still vastly exceeds the 6-fold acceleration threshold compared to manual labor, effectively resolving the temporal bottleneck inherent in traditional analytical laboratories.

\section{Conclusion}

By coupling a controlled, automated brightfield microscopy platform with a specialized Deep Metric Learning pipeline, this work effectively resolves the primary operational bottlenecks inherent to traditional melissopalynology. The proposed system delivers a processing speedup of at least 6x-8x over manual analysis while maintaining expert-level accuracy and providing visually explainable diagnostic features. Empirical evaluations demonstrate that the optimal configuration utilizes a DINOv2 backbone trained via a Multi-Similarity Loss. This architecture, combined with a deterministic preprocessing strategy using a normalized Slate Gray background, efficiently maps targets to a highly separable 128-dimensional unit hypersphere ($\mathbb{S}^{127}$). This approach establishes a highly efficient, scalable, and interpretable framework that outperforms traditional CNN baselines across global latent space metrics.  Future research will expand the scope of these models to include new pollen datasets and human cell analysis. Specifically, we aim to implement automatic morphological classification of red blood cell (RBC) deformations induced by conditions such as sickle cell disease and various anemias \cite{Mohandas2008,Lamoreux2022,Chen2023}. Furthermore, given that erythrocyte mechanical alterations are closely linked to glucose levels and gestational diabetes \cite{Tapia2021, Enriquez2026, Bravo2023}, achieving high-accuracy automated diagnosis remains a critical priority for clinical decision-making.  

\section*{Acknowledgments}
This work is partially supported by grant FONDEF IT24i0064 from Agencia Nacional de Investigación y Desarrollo (ANID) - Chile.

\newpage

\section{Biography Section}

\begin{IEEEbiography}[{\includegraphics[width=1in,height=1.25in,clip,keepaspectratio]{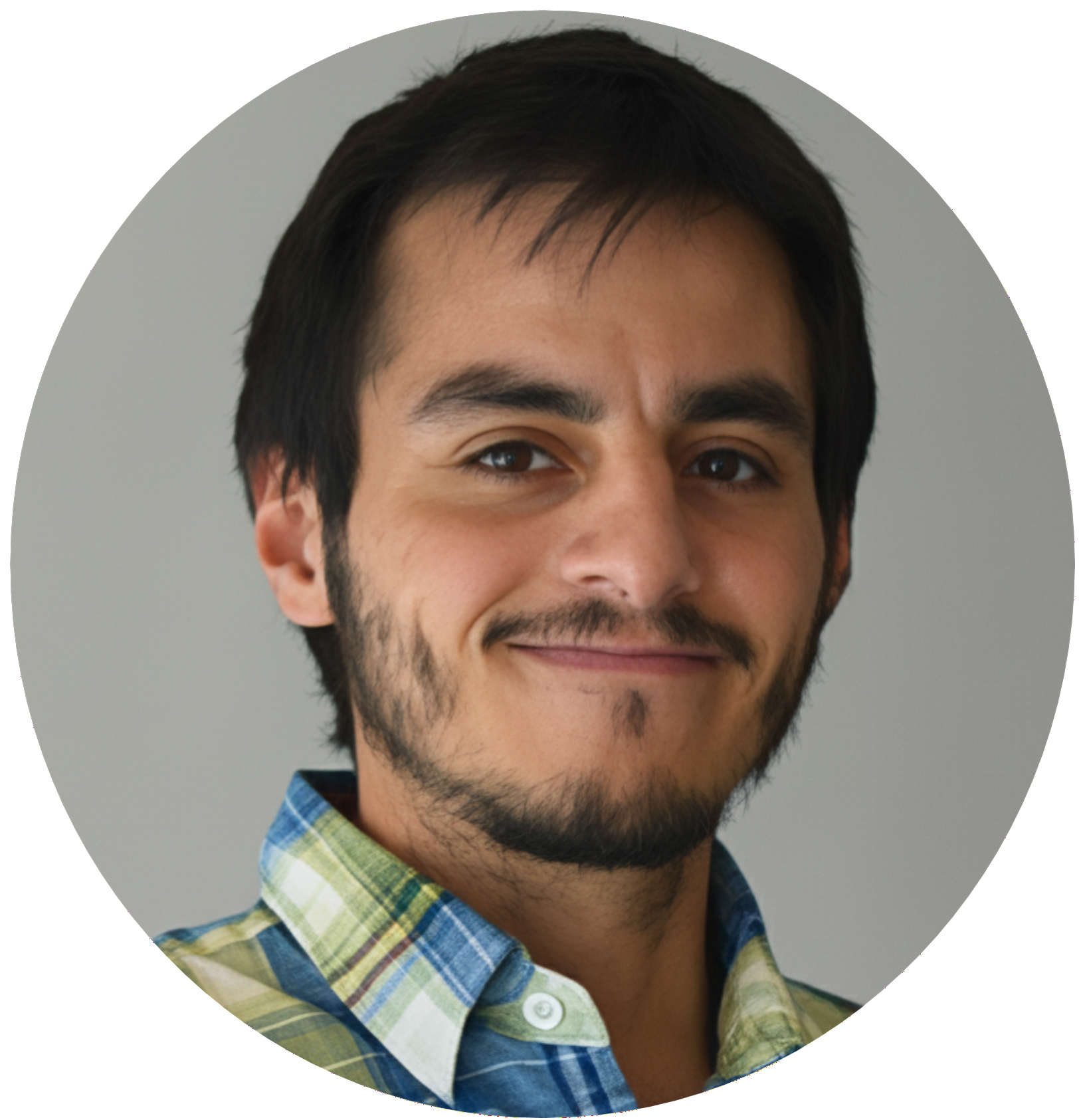}}]{Juan Staforelli-Vivanco, Ph.D.} is a physicist and academic researcher at the Facultad de Ciencias Físicas y Matemáticas of the Universidad de Concepción. As the principal investigator of the Optical Traps and Microscopy Laboratory, his expertise spans photonics, biophysics, and micro-scale systems. His current research focuses on integrating optoelectronic hardware with artificial intelligence, leading applied projects in biological diagnostics and automated agricultural certification. He is highly dedicated to multidisciplinary research, scientific publication, and mentoring new scientists.
\end{IEEEbiography}

\begin{IEEEbiography}[{\includegraphics[width=1in,height=1.25in,clip,keepaspectratio]{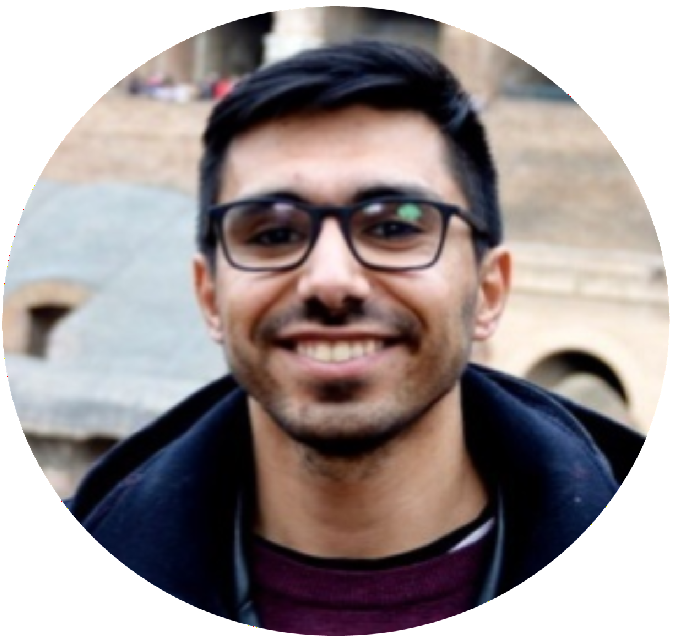}}]{Rodrigo Jofré-Cerda} is an electrical engineer and researcher specializing in automated control in electronics, laser systems and optics, driven by machine/deep learning background. With a Master's degree in Electrical Engineering from the Universidad de Concepción, his work focuses on applying artificial intelligence algorithms to computer vision and laser spectrometry. He has developed on advanced deep learning frameworks such as YOLOv8, U2-Net, and ViTs (DINOv2), for the automated palynological analysis of endemic honeys. His research actively contributes to the modernization of agricultural classification and quality assessment systems.
\end{IEEEbiography}

\begin{IEEEbiography}[{\includegraphics[width=1in,height=1.25in,clip,keepaspectratio]{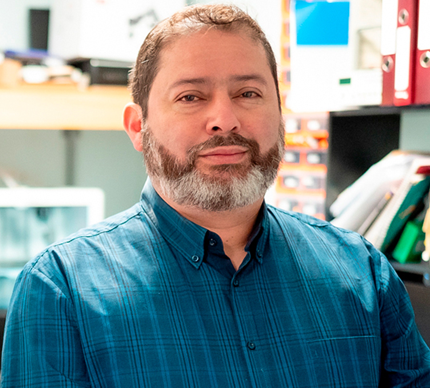}}]{Pablo Coelho, Ph.D.} is an Assistant Professor and researcher at the Universidad San Sebastián, holding a Ph.D. in Electrical Engineering. His primary research focuses on visible and infrared imaging systems, utilizing hyperspectral and microscopic plenoptic technologies. He excels at integrating computer vision and machine learning to solve applied industrial problems in sectors such as mining, beekeeping, and aquaculture. In addition to his research, he teaches undergraduate and graduate courses in physics, prototype design, and digital image.
\end{IEEEbiography}

\begin{IEEEbiography}[{\includegraphics[width=1in,height=1.25in,clip,keepaspectratio]{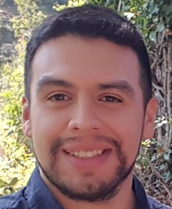}}]{Ignacio Sanhueza, Msc.} is an engineer and researcher affiliated with the Universidad San Sebastián, specializing in artificial intelligence and computer vision. His work primarily focuses on the development and implementation of advanced deep learning architectures, such as YOLO and instance segmentation models. He plays a crucial role in creating high-resolution, specialized datasets for agricultural and environmental research, including automated pollen classification. His multidisciplinary collaborations bridge the gap between complex algorithmic design and practical industrial applications.
\end{IEEEbiography}

\begin{IEEEbiography}[{\includegraphics[width=1in,height=1.25in,clip,keepaspectratio]{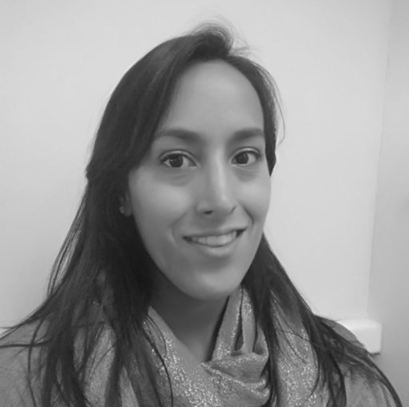}}]{Laura Viafora} is a researcher associated with the Universidad San Sebastián, whose expertise bridges advanced optoelectronics, imaging, and computer vision. Her research contributions include the application of advanced infrared imaging techniques and point spread function estimations for physiological and biomechanical assessments. She also collaborates actively on environmental and agricultural projects, assisting in the development of specialized deep learning datasets for automated microscopic analysis. Her work highlights the effective integration of specialized imaging technologies with modern machine learning frameworks.
\end{IEEEbiography}

\begin{IEEEbiography}[{\includegraphics[width=1in,height=1.25in,clip,keepaspectratio]{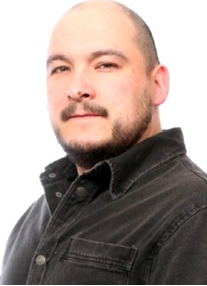}}]{Carlos Toro, Ph.D.} is the Director of the Automation and Robotics Engineering program at the Universidad Andrés Bello (UNAB) in Concepción. His research focuses on optoelectronic sensing, digital signal and image processing, pattern recognition, and machine learning. He actively collaborates on multidisciplinary applied research, leveraging computer vision and artificial intelligence to solve complex agricultural and industrial challenges.
\end{IEEEbiography}

\begin{IEEEbiography}[{\includegraphics[width=1in,height=1.25in,clip,keepaspectratio]{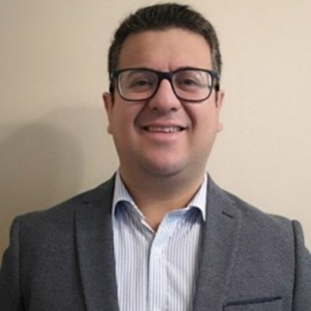}}]{José Max Troncoso} holds a Ph.D. in Environmental Sciences and is a research associate at the Universidad Adventista de Chile and Universidad Arturo Prat. His primary research focuses on palynology, specifically the study of pollen grains and their applications in climate change, environmental science, and value-added agricultural products. He frequently collaborates with interdisciplinary engineering teams to merge botanical sciences with advanced technologies like artificial intelligence. His recent applied work includes using machine learning and hyperspectral microscopy to identify endemic monofloral honeys and detect syrup adulteration. Through this research, he aims to transfer scientific advancements into practical, real-world tools for the apiculture industry.
\end{IEEEbiography}

\begin{IEEEbiography}[{\includegraphics[width=1in,height=1.25in,clip,keepaspectratio]{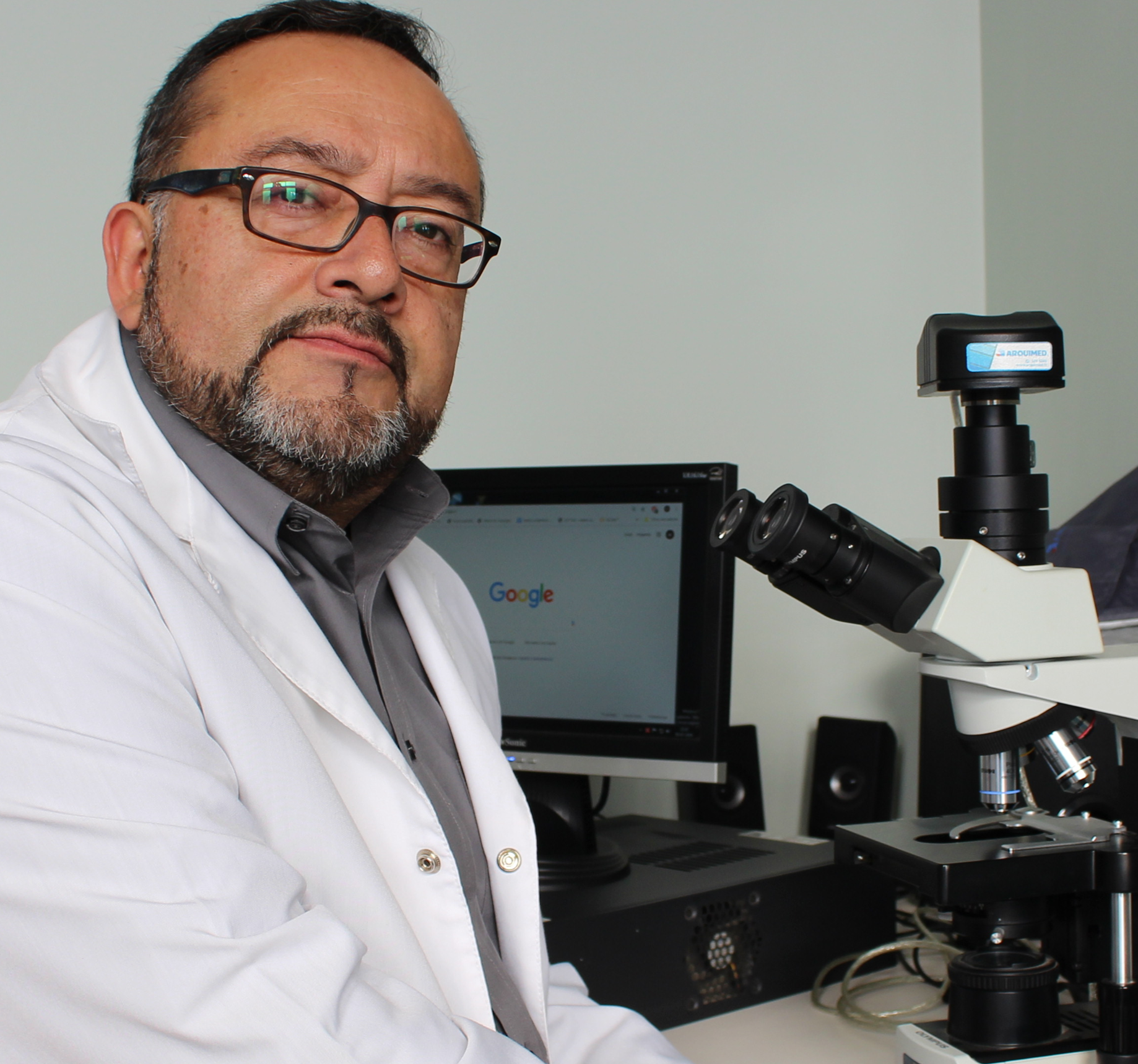}}]{Mauricio Rondanelli, Ph.D.} is a botanist, associate professor, and Vice Principal of the Los Ángeles Campus at the University of Concepción. He holds a Ph.D. in Biological Sciences and serves as Director of the Laboratory of Palynology and Plant Ecology, where he conducts extensive research in Quaternary palynology, melissopalynology, and ecosystem services associated with the forests of south-central Chile. He is also a faculty member of the Center for European Studies at the University of Concepción and a member of the National Apiculture Commission of the Chilean Ministry of Agriculture.

\end{IEEEbiography}

\begin{IEEEbiography}[{\includegraphics[width=1in,height=1.25in,clip,keepaspectratio]{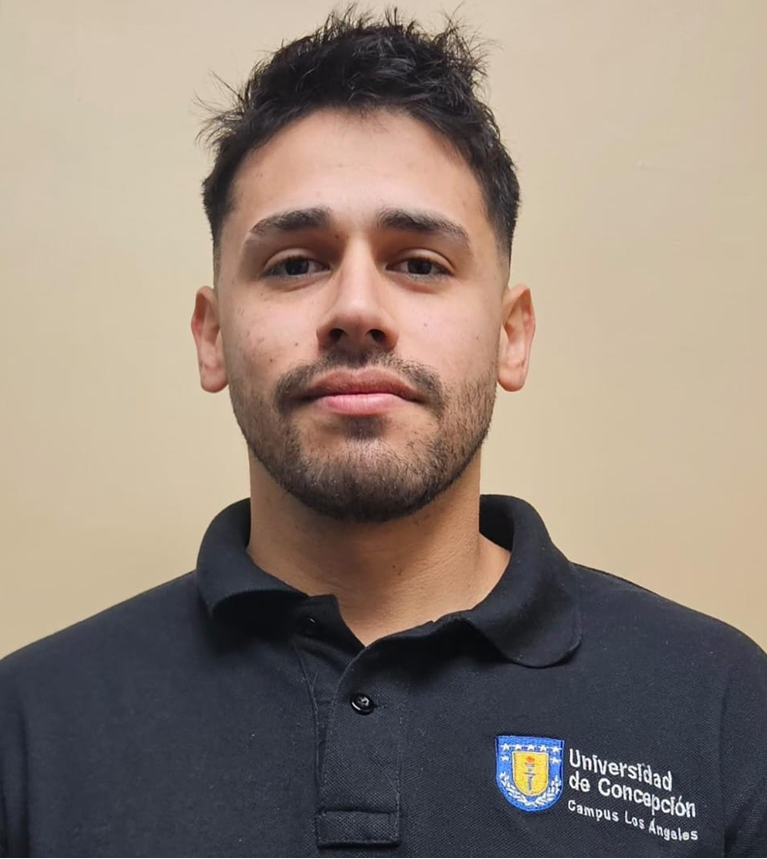}}]{Ivan Lamas} is a Plant Biotechnology Engineer and a professional research assistant at the Palynology and Plant Ecology Laboratory of the Universidad de Concepción, Los Ángeles Campus. Working closely with Dr. Mauricio Rondanelli, he plays a vital role in the palynological analysis of endemic honeys and the management of the university's reference palynotheca. He actively collaborates on multidisciplinary research projects, integrating his biological expertise with engineering teams to detect honey adulteration using artificial intelligence and advanced microscopy. Through his technical and scientific contributions, he actively supports the certification and protection of regional apicultural products.
\end{IEEEbiography}

\begin{IEEEbiography}[{\includegraphics[width=1in,height=1.25in,clip,keepaspectratio]{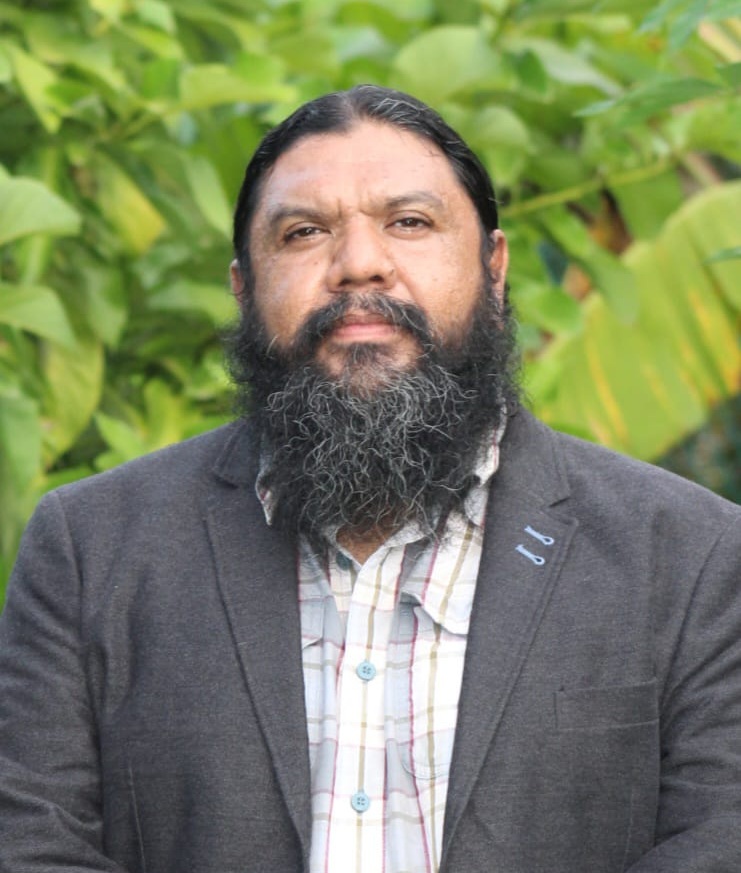}}]{Andy Banegas-Medina, Ph.D.} is a researcher and academic at the Universidad Nacional Autónoma de Honduras (UNAH), affiliated with the Laboratory of Biology at the El Paraíso Campus. Holding a Ph.D. in Environmental Sciences, his expertise encompasses aquatic systems ecology, limnology, and the impacts of climate change. His recent notable research focuses on environmental pollution, specifically the identification of potentially toxic trace elements within the marine ecosystems of the Cayos Cochinos archipelago. He actively fosters international academic cooperation, frequently collaborating with institutions such as the Universidad de Concepción in Chile. Through his scientific contributions, he aims to advance conservation planning and protect highly vulnerable natural resources.
\end{IEEEbiography}

\begin{IEEEbiography}[{\includegraphics[width=1in,height=1.25in,clip,keepaspectratio]{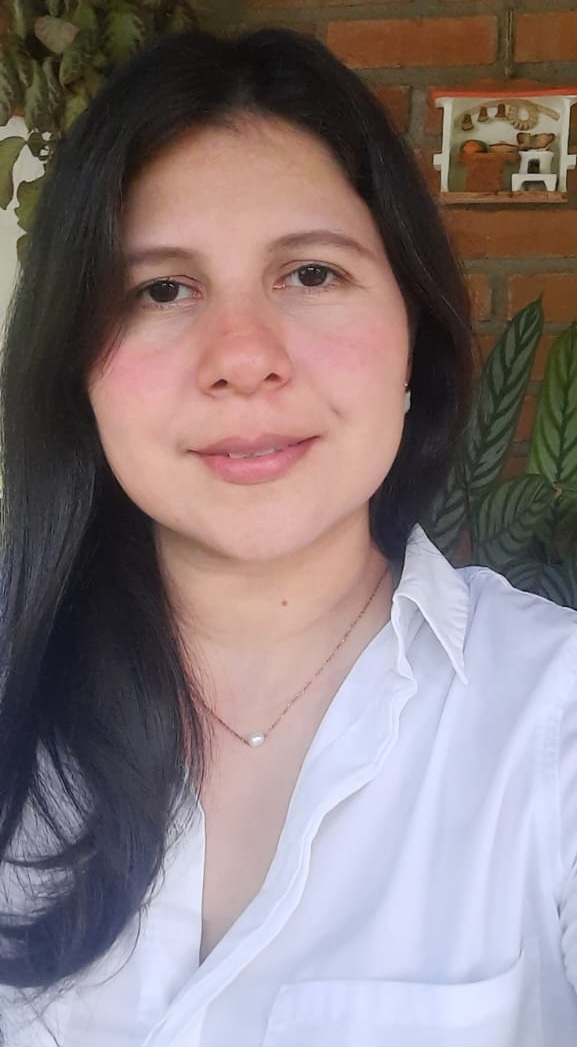}}]{Isis-Yelena Montes, Ph.D.} is an academic researcher at the Laboratory of Biology within the Universidad Nacional Autónoma de Honduras (UNAH), El Paraíso Campus. Her research spans aquatic systems, environmental science, and applied biogeochemistry, with a strong commitment to sustainable development. She frequently collaborates with international institutions, including the Universidad de Concepción in Chile, to study trace element pollution in coastal environments. Notably, her recent fieldwork investigates the presence of potentially toxic elements within the fragile marine ecosystems of the Cayos Cochinos archipelago. Her scientific contributions are vital for guiding ecological risk assessments and shaping conservation policies for vulnerable habitats.
\end{IEEEbiography}

\begin{IEEEbiography}[{\includegraphics[width=1in,height=1.25in,clip,keepaspectratio]{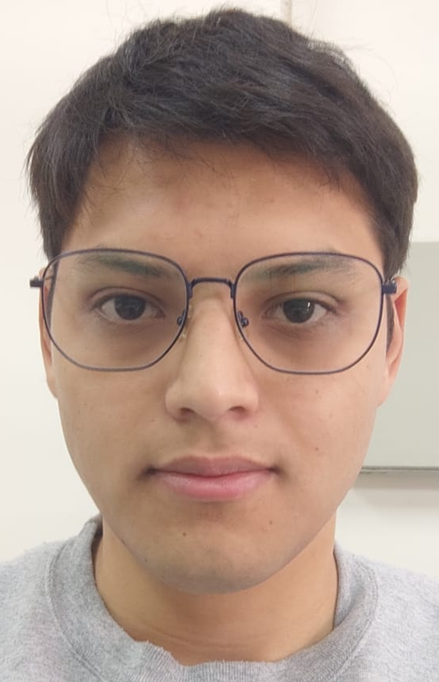}}]{Benjamín Muñoz} is a physics and engineering physics student at the Universidad de Concepción, currently developing his professional title thesis under the guidance of Dr. Juan Pablo Staforelli. His research investigates the use of pollen concentrates as a metric for detecting adulteration in endemic honeys, employing optoelectronic microscopy systems and neural networks. He is an active member of a multidisciplinary research team, co-authoring papers on automated palynological analysis using deep metric learning and U2-Net architectures. Furthermore, he has contributed to the creation of specialized, high-resolution datasets for pollen grain instance segmentation.
\end{IEEEbiography}

\begin{IEEEbiography}[{\includegraphics[width=1in,height=1.25in,clip,keepaspectratio]{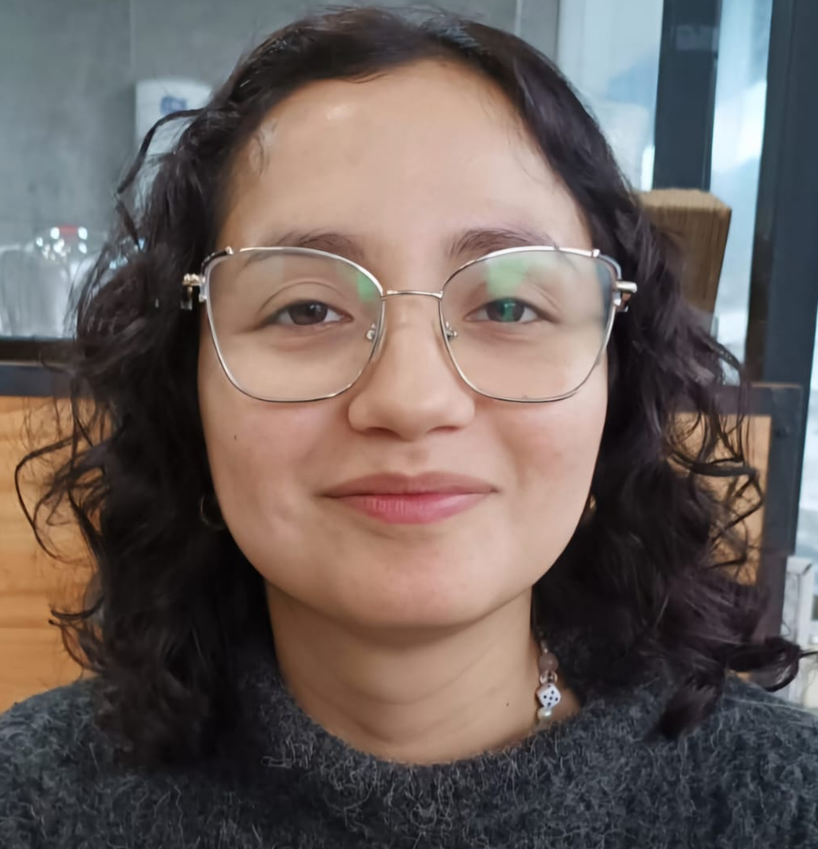}}]{Valentina Salamanca, Bs.c.} is a physics and engineering physics student at the Universidad de Concepción, currently developing her professional title thesis under the mentorship of Dr. Juan Pablo Staforelli. Her specialized research focuses on validating reference volumetry in brightfield microscopy to calibrate pollen reconstructions using digital holography. She is an active contributor to multidisciplinary optoelectronics projects, applying her skills to advanced computational imaging techniques. Furthermore, she has co-authored scientific publications and high-resolution datasets concerning the 3D volumetric reconstruction of pollen grains.
\end{IEEEbiography}

\begin{IEEEbiography}[{\includegraphics[width=1in,height=1.25in,clip,keepaspectratio]{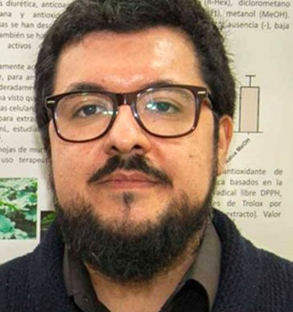}}]{Marcelo González-Ortiz, Ph.D} is an Associate Professor in the Department of Obstetrics and Gynecology at the Universidad de Concepción, holding a Ph.D. in Physiological Sciences. As the Principal Investigator of the Maternal-Fetal Research Laboratory (LIMaF), he specializes in vascular physiology and the cellular mechanisms of pregnancy pathologies like gestational diabetes. His work bridges clinical obstetrics with biophysics, frequently collaborating on multidisciplinary projects to assess mother-child health outcomes. Notably, he utilizes advanced optical techniques, such as membrane fluctuation spectroscopy, to study the micro-rheological properties of erythrocytes under pathological stress. Through his extensive research and publications, he actively advances the field of new methods in microscopy to biomedicine.
\end{IEEEbiography}

\begin{IEEEbiography}[{\includegraphics[width=1in,height=1.25in,clip,keepaspectratio]{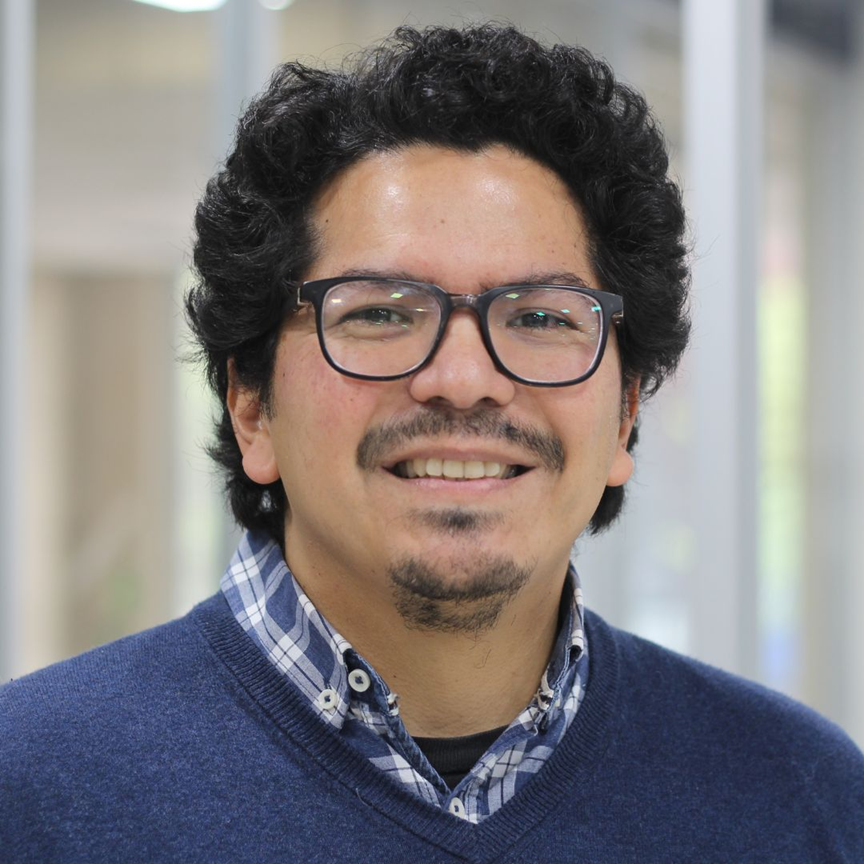}}]{Esteban Vera, Ph.D.} is an Adjunct Professor in the School of Electrical Engineering at the Pontificia Universidad Católica de Valparaíso (PUCV). He serves as the founder and director of the Optoelectronics Laboratory (Optolab), focusing his research on optoelectronics, computational imaging, and image sensing. Holding a Ph.D. in Engineering Sciences, he develops advanced technologies like AI-driven wavefront sensors for satellite communications and adaptive optics. He frequently collaborates on multidisciplinary projects, including neuromorphic imaging, to advance space surveillance and high-speed optical connectivity.

\end{IEEEbiography}

\vfill

\end{document}